%% file: main.tex
\documentclass[preprint,12pt]{elsarticle}
\input{packages}

\usepackage{amssymb}

\usepackage{comment}
\usepackage{multirow}
\usepackage{adjustbox}
\usepackage{arydshln}

\usepackage{float}
\floatstyle{plaintop}
\restylefloat{table}

\journal{International Journal of Information Management Data Insights}

\begin{document}

\begin{frontmatter}

\title{Learning Transactions Representations for Information Management in Banks: Mastering Local, Global, and External Knowledge}


\affiliation[label1]{organization={Skolkovo Institute of Science and Technology (Skoltech)},
            addressline={Bolshoy Boulevard, 30 p.1},
            city={Moscow},
            postcode={121205},
            state={},
            country={Russia}}
\affiliation[label2]{organization={Sber AI Lab},
            addressline={Kutuzovsky Avenue, 32},
            city={Moscow},
            postcode={121165},
            state={},
            country={Russia}}
\fntext[cor1]{Equal contribution}
\cortext[cor2]{Corresponding Author}
\author[label1]{Alexandra Bazarova\fnref{cor1}}
\author[label1]{Maria Kovaleva\fnref{cor1}}
\author[label1]{Ilya Kuleshov\fnref{cor1}\corref{cor2}}
\ead{i.kuleshov@skoltech.ru}
\author[label1]{Evgenia Romanenkova\fnref{cor1}}
\author[label1]{Alexander Stepikin\fnref{cor1}}
\author[label1]{Aleksandr Yugay\fnref{cor1}}
\author[label2]{Dzhambulat Mollaev}
\author[label2]{Ivan Kireev}
\author[label2]{Andrey Savchenko}
\author[label1]{Alexey Zaytsev}

\begin{abstract}

In today's world, banks use artificial intelligence to optimize diverse business processes, aiming to improve customer experience. Most of the customer-related tasks can be categorized into two groups: 1) local ones, which focus on a client's current state, such as transaction forecasting, and 2) global ones, which consider the general customer behaviour, e.g., predicting successful loan repayment. Unfortunately, maintaining separate models for each task is costly. Therefore, to better facilitate information management, we compared eight state-of-the-art unsupervised methods on 11 tasks in search for a one-size-fits-all solution. Contrastive self-supervised learning methods were demonstrated to excel at global problems, while generative techniques were superior at local tasks. We also introduced a novel approach, which enriches the client's representation by incorporating external information gathered from other clients. Our method outperforms classical models, boosting accuracy by up to 20\%.
\end{abstract}



\end{frontmatter}


\section{Introduction}
\label{chap:intro}
\input{1-intro}

\section{Related works}
\label{chap:review}
\input{2-review}

\section{Methodology}
\label{chap:method}
\input{3-methodology}

\section{Models}
\label{chap:models}
\input{4-models}

\section{Data overview}
\label{chap:data}
\input{5-data}

\section{Results}
\input{6-results}

\section{Conclusions}
\input{7-conclusions}

\section{Acknowledgments}
The research was supported by the Russian Science Foundation grant 20-7110135.


\appendix
\input{8-appendix}



\clearpage
\bibliographystyle{elsarticle-num}
\bibliography{ref}
\end{document}

%% file: packages.tex
\usepackage{graphicx}
\usepackage{subcaption}
\usepackage[dvipsnames]{xcolor}
\usepackage{marvosym}
\usepackage{multirow}
\usepackage{hyperref}
\usepackage{amsmath}
\usepackage{floatrow}

\newfloatcommand{capbtabbox}{table}[][\FBwidth]

%% file: 1-intro.tex
The banking industry must keep up with current trends to remain competitive.
Technological advancements such as Artificial Intelligence (AI) have significantly transformed how banks process and analyze data, allowing them to extract valuable insights for more informed decision-making and effectively address complex challenges~\cite{BANYMOHAMMED2024100215,BUENO2024100230}.

Among various types of data collected and managed by banks, AI-based analysis of financial transactions appears to be the most promising for various operational and strategic purposes~\cite{AMATO2024100234, kaya2019artificial}. 
Such solutions already play a crucial role in risk assessment, improving the accuracy of credit scoring, advanced bankruptcy prediction, and financial distress identification~\cite{SINGH2022100094}.
They also help to automate fraud detection, identify transaction anomalies, and protect both banks and their customers~\cite{jurgovsky_sequence_2018}. 
Moreover, AI-driven customer segmentation and personalized marketing strategies allow banks to tailor services to individual customer needs, enhancing customer satisfaction and loyalty~\cite{AMATO2024100234}. 
While these approaches enable banks to optimize operations, reduce costs, and manage risks, they often depend on task-specific methodologies, large labeled datasets, and expert-defined features, limiting their usage.

All these challenges may be addressed by representation learning via neural networks, a powerful solution that benefits from the vast amounts of transaction history sequences stored by modern banks. 
It focuses on extracting hidden patterns from data, capturing intrinsic features into low-dimensional embeddings. 
These representations can further be utilized in various downstream tasks, offering a more scalable and adaptable approach to employing data for business-oriented goals~\cite{babaev2022coles, bin2022review,  li2023new, mancisidor2021learning, romanenkova2022similarity}.
This general pipeline of SSL applications in the banking industry is further discussed in Section \ref{chap:method}.

From a business perspective, representations of the customers' transaction sequences are supposed to reflect data properties at both global and local levels.
We define \textit{global properties} as a characteristic of a customer's transaction history as a whole.
In this case, high-quality global representations should be helpful when comparing and classifying the sequences as entities~\cite{bin2022review, jaiswal2020survey}.
In contrast, the state of a customer at a specific point in time is described by \textit{local properties}. 
Local representations are used for tasks associated with momentary customer behavior, such as next-event prediction~\cite{zhuzhel2023continuous} or credit card fraud detection~\cite{heryadi_learning_2017}. 
Moreover, local information should reflect the \textit{dynamics} of clients' lives and be useful for change point detection~\cite{ala2022deep, deldari2021time}. 
However, the methods in the literature typically focus on either global or local embedding properties, completely ignoring representation dynamics.


Another drawback of existing models is that they do not consider the external context when creating an embedding for each customer. 
While choosing macroeconomic indicators requires expert knowledge, such information may be represented by the common characteristics of the other customers' transactions, which were shown to be strongly correlated~\cite{begicheva2021bank, thomas2000survey}. 
Thus, a reasonable approach is to use different aggregations of the representations of all customers to form the external context vector. 
Previous works ignored this concept or designed very narrow methods, significantly increasing the number of the model parameters size~\cite{GomezRodriguez2011UncoveringTT}. 
However, accounting for this information may potentially boost the quality of the models~\cite{ala2022deep}.

To conclude, this work develops a representation learning method that considers local and global data patterns in transactional data. 
Additionally, the paper explores ways of incorporating the external context into the obtained representations. 
To solve these tasks, we extend existing strategies of working with sequential data and adopt the most promising techniques. 
More particular, the main contributions of the paper are the following: 

\begin{enumerate}
    \item We propose several techniques based on generative self-supervised learning to obtain transactional data representations with strong local and global properties. Our methods offer a good trade-off between these properties across various scenarios, providing the most versatile representations for transaction data at the moment;
    \item We suggest an efficient procedure that utilizes the attention mechanism to involve the external context when constructing data representations, further enhancing the model quality for most applied problems;
    \item We provide an extensive pipeline for evaluating the informativeness of data representations in terms of domain-specific properties that suit various needs.  Our pipeline includes four diverse, open datasets, four downstream problems for them, and one common task. We also validate the models on an industrial-scale private dataset. The source code to run our benchmark and reproduce all experiments is publicly available\footnote{\url{https://github.com/romanenkova95/transactions_gen_models}}.
\end{enumerate}
  
The rest of the article is organized as follows. 
The next section discusses related research on representation learning in the transactional data domain. 
Section 3 describes the proposed methods and the datasets used. 
In Sections 4 and 5, we present the results of our experiments, which evaluated the global and local properties of the data. 
Section 6 contains our approach to considering the external context. 
Sections 6 and 7 are dedicated to discussing our results and conclusions. 

%% file: 2-review.tex

This section briefly describes the existing self-supervised representation learning methods for transactional data.
Generally, there are two SSL pa\-ra\-digms: contrastive and generative.
Below, we describe the most relevant models among these two approaches.
We start the section by mentioning the supervised models relevant to the banking domain.
Finally, we accompany the review with the Temporal Point Process (TPP) modeling methods explicitly designed for the analysis of event sequences --- the type of data transaction histories belong to.

\subsection{Supervised methods for representation learning}
\label{subsec:review_supervised_methods}
Solutions in banking often lean towards traditional paradigms incorporating expert domain knowledge~\cite{AMATO2024100234}. 
In credit scoring, researchers prefer the classical machine learning methods~\cite{dastile2020statistical} since they tend to outperform deep learning approaches on top of manually constructed features~\cite{gunnarsson2021deep, marceau2019comparison, moscato2021benchmark, schmitt2022deep}. 
Credit card fraud detection methods analysis provides similar insights~\cite{bahnsen2016feature,jurgovsky_sequence_2018,zaytsev2023designing,zhang2021hoba}.

On the other hand, recent papers support applying supervised deep learning methods to raw transactional data, modifying approaches to account for the particularities of the target domain. 
Thus, when working with sequential data, modern works consider recurrent neural networks (RNNs) and Transformers~\cite{ala2022deep, babaev2019rnn,  wang2018deep, wang2022deep}. 
Other notable solutions include adjusting for the time since the last transaction~\cite{xie_learning_2022} and working with small subsequences to curb the vanishing gradients problem~\cite{forough_ensemble_2021}.

\subsection{Self-Supervised Learning Paradigm}
\label{subsec:review_ssl_methods}
Lately, there has been a surge in attention to self-supervised methods (SSL).
Such methods try to extract information from unlabeled data. 
This allows researchers to skip the costly gathering of expert annotations, leveraging the large bodies of low-cost raw data. The SSL methods can be roughly divided into contrastive and generative approaches~\cite{zhang2020self}.

\emph{Contrastive approaches} have been gaining popularity in financial transactions~\cite{babaev2022coles, li2023new, li_deep_2020}. 
All contrastive methods aim to yield close representations of ``similar'' objects and distant representations of ``dissimilar'' ones. 
The way one measures similarity may be very simple, such as the one in the papers~\cite{babaev2022coles, moskvoretskii2024self, romanenkova2022similarity}, which contrasts between slices from different sequences.
More complex options include combining distance and angle-based metrics~\cite{li_deep_2020} and training with hierarchical losses~\cite{yue2022ts2vec}.

\emph{Generative models} use various training approaches to learn the hidden data distribution. 
The resulting knowledge may then be utilized to produce plausible-looking data. 
These approaches often originate from Neural Language Processing (NLP)~\cite{kenton2019bert, radford2019language}.
Autoencoder is a very popular generative method due to its simplicity and efficiency in different modalities~\cite{tschannen2018recent}. 
The article~\cite{mancisidor2021learning} employs the variational autoencoder to learn bank customer representations, which is useful for downstream tasks~\cite{SINGH2022100094}. 
Masked language models (MLMs) take the autoencoder idea one step further. 
Such models are trained via restoring randomly changed tokens and perform well on diverse benchmarks~\cite{kenton2019bert, he2022masked}.

According to the latest research, generative methods outperform contrastive ones at missing value prediction, among other tasks~\cite{jaiswal2020survey}. 
This fact may be indicative of better local properties. 
On the other hand, while supposedly good locally, generative models are known to be less efficient at global tasks.
Here, contrastive models, like CoLES~\cite{babaev2022coles}, come into play because they typically consider a sequence of events as a whole, enforcing the embeddings' global properties.
As no evident leader exists, we explore both approaches for transactional data, testing the resulting representations for the desired properties. 

\subsection{Temporal Point Process Models}
\label{subsec:review_tpp}
Financial transactional data can be treated as event sequences, differing from conventional time series and natural language data in some key aspects:

\begin{itemize}
   \item Observations in time series appear uniformly in time, which is not valid for transactional sequences~\cite{moskvoretskii2024self}. In turn, text is not tied to time at all.
   \item In contrast to time series datasets, the sequences in a transactional dataset may be of different lengths.
   \item Transactions have heterogeneous features, e.g., MCC codes are categorical, and transaction amounts are continuous. Generally, neither the time series nor the NLP data exhibit this peculiarity.
\end{itemize}

Event sequence problems are commonly solved using TPP models.
Following this approach, one generally aims to reconstruct conditional event intensity, determining the type and frequency of future events based on the history of earlier observations~\cite{hawkes1971spectra, liniger2009multivariate}. Standard objectives include likelihood maximization, time-to-next-event (return time) estimation, and next-event-type prediction~\cite{mei2017neural, zhuzhel2023continuous}.
 
Modern research incorporates neural networks for the intensity function parametrization.
For example, the Neural Hawkes Process (NHP)~\cite{mei2017neural} applies the ideas of the classic Hawkes Process~\cite{hawkes1971spectra} to RNN-generated latent embeddings. 
The methods from papers~\cite{mei2021transformer, zhang2020self, zuo2020transformer} are based on the same idea but stick to the architectures with the attention mechanism~\cite{vaswani2017attention}.
Another class of neural TPP models includes recently proposed continuous convolutional networks (CCNNs)~\cite{li2020learning, shi2021continuous}.
This architecture successfully handles the irregularity of event sequences in many cases.
For example, the COTIC~\cite{zhuzhel2023continuous} approach uses a deep CCNN backbone model and two multilayer perceptron heads on top of it for return time and event type prediction.

Note that TPP models are not typically studied in terms of their embedding quality, which is the goal of this research.
However, the models' natural problem statement~\cite{mei2017neural, zhuzhel2023continuous} may enhance highly efficient representations of event sequences regarding their local properties.
Consequently, we add several TPP models to our benchmarking.

%% file: 3-methodology.tex
\begin{figure}[!ht]
    \centering
    \includegraphics[width=\textwidth]{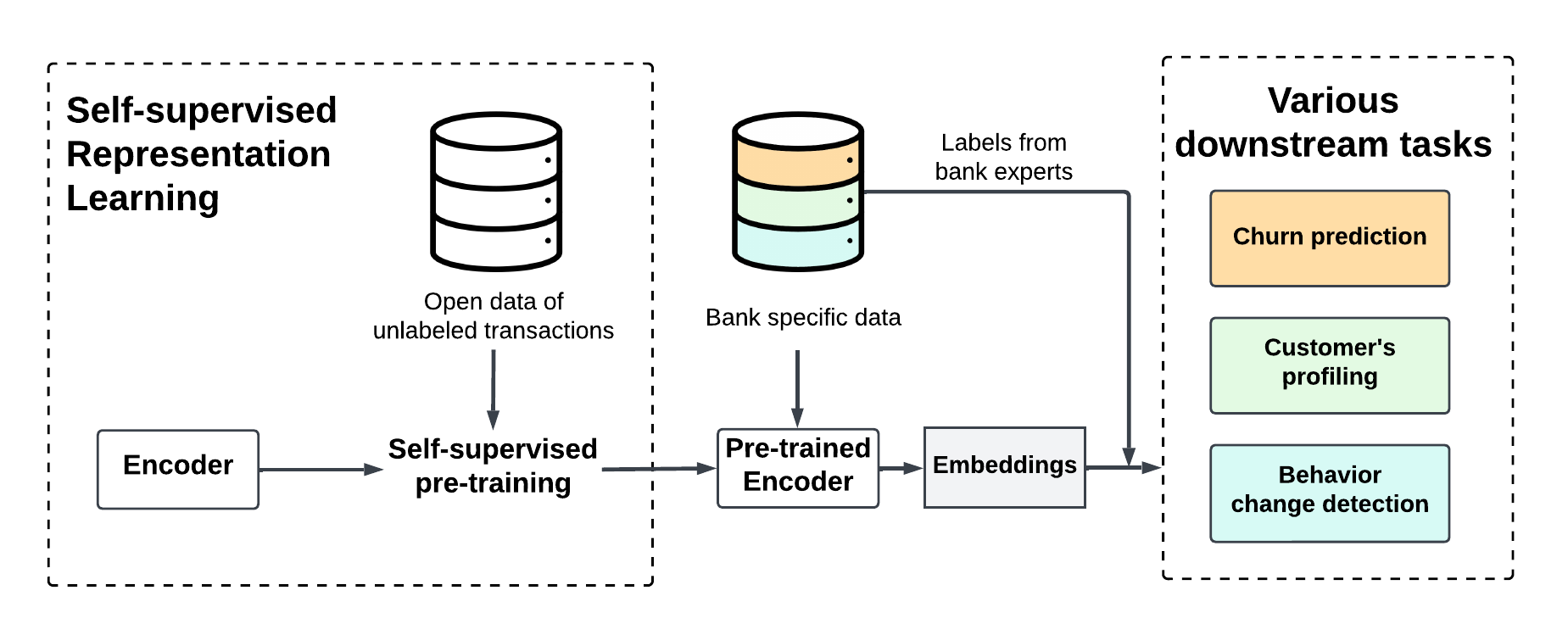}
    \caption{The general flow of the application of self-supervised learning to industrial tasks in banking.}
    \label{fig:flow-chart}
\end{figure}

\label{sec:method}
The general pipeline for incorporating self-supervised learning in the banking industry is illustrated by Figure~\ref{fig:flow-chart}.
The general idea is to pre-train an \emph{encoder model} on large open corpuses of unlabelled data, producing informative embeddings for transactional sequences.
These representations can then be used for solving downstream tasks.
Further information on the specific approaches to learning such embeddings can be found in Section~\ref{chap:models}. 
In the remainder of this section, we describe our approach to evaluating the resulting representations.

As outlined in Section~\ref{chap:intro}, different business tasks in banking often rely on different aspects of embeddings.
In this work, we study the trade-off between local and global properties of transactional data representations obtained using various models.
\emph{Global properties} characterize the client's behaviour throughout the entire history of his/her transactions. 
In contrast, \emph{local properties} show the nature of the client's current state and its evolution through time.

\subsection{Global properties' validation methods}
\label{subsec:global_validation}
To assess the quality of the obtained representations regarding their global properties, we follow the paper~\cite{babaev2022coles}, which proposed the CoLES method. 
More precisely, following the approach illustrated in Figure~\ref{fig:flow-chart}, we evaluate the obtained embeddings on several downstream tasks.
This requires each dataset used in this work to have a classification problem behind it.
For example, the \emph{Churn} dataset contains binary targets: for each bank customer, we aim to identify whether he left the bank (see Section~\ref{chap:data} for complete information about considered data). 

To sum up, the global evaluation pipeline consists of three steps:
\begin{enumerate}
    \item For an initial sequence of transactions of length $T_{i}$ related to the $i$-th user, we obtain the \emph{global representation} $\boldsymbol{H}^{i}\in\mathbb{R}^{d}$, which characterizes the transaction history as a whole.
    In the case of sequence-to-sequence encoder architectures, a pooling operation is applied to the encoded sequence to create a single representation of an entire series.
    
    \item Given representation $\boldsymbol{H}^{i}$, we predict the target label $y_{i}\in\{0, 1, \ldots, C\}$ using gradient boosting. 
    We stick to the LightGBM~\cite{ke2017lightgbm} model applied in the paper~\cite{babaev2022coles}. 
    It works fast enough for large data samples and provides sufficiently high-quality results. The hyperparameters of the boosting model are fixed and do not vary across all the base models under study.

    \item Classification results are evaluated using a standard set of metrics described in~\ref{sec:metrics}.
\end{enumerate}
 
\subsection{Local properties validation methods}
\label{subsec:local_validation}
We suggest using a set of techniques to evaluate the quality of the obtained representations in terms of their local properties. 

\paragraph{\textbf{Local downstream targets}}
By analogy with the global downstream tasks discussed above, business problems frequently require local clients' information to solve them.  
We propose considering such local downstream classification tasks for the Churn, Default, and HSBC datasets (see Section~\ref{chap:data}).

In the \emph{Churn} dataset, original target labels indicate clients who eventually stopped using the bank's services.
It is reasonable to assume that the behaviour of customers who are about to leave a bank begins to change in advance. 
For example, they make fewer transactions and stop topping up their card balances.
Given this, all transactions by the "churn" user $i$ within a month before his/her departure are tagged with $c_j^i=1$, while the others are marked with $c_j^i = 0$. 
We select an "early outflow" horizon of one month for empirical reasons.

The \emph{Default} dataset also follows a similar logic. Here, we create local binary labels that reflect the client's transition to the "pre-default" state, noting that the default of a client typically corresponds to three missing monthly instalment payments.

\emph{HSBC} initially contains local binary labels for each transaction record, indicating whether it is malicious.
In the case of this dataset, we do the opposite and add global targets that show if a client has ever been a victim of bank fraud.

As a result, we propose to evaluate the quality of local representations by predicting a local binary target $c^{i}_{j}$. 
The corresponding classification tasks are solved using a two-layer perceptron. 
The better local embedding describes the client's current behaviour, the better a simple classifier model can solve applied problems.

\paragraph{\textbf{Next transaction MCC prediction}}
Next-event-type prediction is a common objective in TPP modelling~\cite{mei2017neural, mei2021transformer, zhuzhel2023continuous}.
Formally, it implies solving the multiclass classification problem: given a "frozen" local representation $\boldsymbol{h}_{t_j}^{i}$, a separate model predicts the following MCC (event type) for the $i$-th client.
To achieve this, we train an additional, simple two-layer perceptron.

It is important to consider that there are a lot of rare MCCs in the datasets. 
From a business perspective, such categories are often less meaningful. 
Therefore, to simplify the task, we kept only the transactions with the top-$100$ most popular codes.

\paragraph{\textbf{Challenge in inferring local representations}}
Generally, not all the studied models are efficient in a sequence-to-sequence mode. 
That is, they may be unable to produce representations for a specific timestamp in a given sequence. 
Thus, we use a sliding window of size $w$ to obtain the local representations in all cases.

In particular, for the $i$-th user at time $t_{j}\in[t_{w}, T_{i}]$, we consider the subsequence $\boldsymbol{S}_{j-w:j}^{i }$. Here, $T_i$ denotes the length of transactional history for this client.  
Next, the interval $\boldsymbol{S}_{j-w:j}^{i }$ is passed through the encoder model to obtain a \emph{local representation} $\boldsymbol{h}_{t_j}^{i}\in\mathbb{R}^{d}$.
This approach is illustrated in Figure~\ref{fig:local_validation_coles}.
\begin{figure}[!ht]
     \centering
     \includegraphics[width=0.75\columnwidth]{local_val.pdf}
     \caption{Scheme of the sliding window approach for local validation.}
     \label{fig:local_validation_coles}
\end{figure}

On the one hand, this technique helps us create local embeddings for any given model. 
Moreover, it enhances local representation properties, as we do not consider out-of-date information. 
On the other hand, a tiny window size may be insufficient to capture the broader patterns of a customer at a given time step. 
Additionally, the process does not permit obtaining local embeddings for times $t \leq t_{w}$. 
However, this issue should be relatively insignificant in the case of long transaction sequences. 
Therefore, $w$ becomes an additional hyperparameter that must be tuned to balance the above trade-off.

\subsection{Strategy for validation of embeddings dynamism}\label{sec:cp_methods}

An alternative approach for assessing the quality of the local representations is to evaluate their \emph{dynamic properties}, i.e., the ability to reflect changes in the raw data.
In this regard, we considered the change point detection (CPD) task~\cite{van2020evaluation, romanenkova2022indid, Truong_2020} posed in the space of local representations.

Change points are time moments with an abrupt shift in sequential data distribution that may indicate transitions between states in the underlying process. 
If we consider the behaviour of bank customers as such a process, an example of a change point could be taking a loan or getting a new job. 
The task of change point detection is to identify the moments of such shifts accurately.

Let $\boldsymbol{H} = \{\boldsymbol{h}_{t_j}\}_{j = 1}^N \subset \mathcal{H} \subset \mathbb{R}^d$ be the sequence of the client's local representations $\boldsymbol{h}_{t_j}$, where $t_j$ is the timestamp of $j$-th transaction, $t_1 \leq \dots \leq t_N$. It is natural to assume that if there is a change point in a time series at a particular moment $\tau$, then, for any $t_i < t_j < \tau < t_k$, the following holds with high probability:
\[
d(\boldsymbol{h}_{t_i}, \boldsymbol{h}_{t_j}) < d(\boldsymbol{h}_{t_i}, \boldsymbol{h}_{t_k}),
\]
where $d(\cdot, \cdot)$ is a metric assigned to $\mathcal{H}$. Following ~\cite{deldari2021time, romanenkova2022indid}, we select the cosine similarity as our distance measure $d$. This metric is used in a series of experiments to illustrate how the representations evolve in the vicinity of a change point. To identify the change points in the embedding sequences, we employ the standard Dynp~\cite{Truong_2020} method from the ruptures package\footnote{\url{https://centre-borelli.github.io/ruptures-docs/}}.

\subsection{Validation metrics}
\label{sec:metrics}

We utilize a set of classical metrics in our benchmark. As mentioned above, our validation techniques involve binary and multiclass classification problems, as well as change point detection tasks.

\paragraph{\textbf{Classification metrics}}
For classification problems, we utilize the standard metrics: Accuracy, ROC-AUC, and PR-AUC~\cite{ marceau2019comparison, moscato2021benchmark, romanenkova2022similarity}. 
For these three metrics, a better model produces higher values.

Accuracy is defined as the ratio of correct predictions to all predictions made:
\begin{equation}    
    \text{Accuracy}=\frac{TP + TN}{TP + TN + FP + FN}, \label{eq:accuracy_def}
\end{equation}
where TP --- true positive, TN --- true negative, FP --- false positive, and FN --- false negative predictions.  

ROC-AUC, or the area under the Receiver Operating Characteristic (ROC) curve, represents the relationship between true positive rate (TPR or Recall) and false positive rate (FPR) for different thresholds. Similarly, the PR-AUC is the area under the Precision-Recall curve, which describes the connection between Precision and Recall (TPR). Formally, 
\begin{equation}
    \text{TPR} = \frac{TP}{TP + FN}, \quad
    \text{FPR} = \frac{FP}{TP + FN}, \quad
    \text{Precision} = \frac{TP}{TP + FP}.
\end{equation}

To generalize to multiple classes, we use micro-averaging for Accuracy and weighted averaging for ROC-AUC and PR-AUC. 
This means that the contribution of each class to the overall result is proportional to the number of its representatives. 
Consequently, the result is insensitive to class imbalance. 
 
\paragraph{\textbf{Change point detection quality}}
We also evaluate the dynamics of the obtained embeddings via change point detection methods.
Sticking to this field's best practices, we use the following CPD quality metrics: average detection delay (DD)~\cite{romanenkova2022indid} and detection accuracy~\cite{van2020evaluation} with different values of the margin parameter $m$. 

DD is the average difference between the estimated and actual change points:
\begin{equation}
\mathrm{DD} = \frac{1}{T} \sum_{i=1}^T (\hat{\tau}_i - \tau_i) I[\hat{\tau}_i \geq \tau_i],
\end{equation}
where $\tau_i$ is the true change point, $\hat{\tau}_i$ is the predicted one, $T$ is the total number of disorders and $I[\cdot]$ is the indicator function. 

We assume that there is only one disruption per sequence. 
The detection accuracy $\mathrm{A}_m$ determines how often the detected change falls within the $m$-the neighbourhood of the true change point: 
\begin{equation}
\mathrm{A}_m = \frac{1}{T} \sum_{i=1}^T I[|\hat{\tau}_i - \tau_i| \leq m ].
\end{equation}

An efficient change point detector has low detection delay and high detection accuracy for different margins $m$.

%% file: 4-models.tex
Below, we describe specific encoder models and their modifications: supervised baselines in subsection~\ref{sec:supervised}, contrastive methods CoLES and TS2Vec in subsection~\ref{sec:contrastive_methods}, generative models such as autoencoders and autoregressive model in subsection~\ref{sec:ae}, and TPP models in subsection~\ref{sec:methods_tpp}. 
Finally, subsection~\ref{sec:global_context_methods} concludes with a description of our novel methods for obtaining and using the external context.

\subsection{Supervised baselines}
\label{sec:supervised}

Obtaining an accurate markup for a real-world dataset is expensive and time-consuming. Nevertheless, supervised methods remain popular first-step solutions for representation learning problems~\cite{dastile2020statistical,jurgovsky_sequence_2018,zhang2021hoba}. We, in turn, focus on self-supervised methods in our work, renowned for their capacity to uncover latent data patterns from unlabeled data. To provide a deep understanding of the possible gap between embeddings from these approaches, we train several baseline models that adopt convolutional and recurrent neural network architectures. 

In the experiments, we show the results for the model with the best performance on downstream tasks (see Section~\ref{chap:method}).
The selected architecture consists of a single-layer Gated Recurrent Unit (GRU)~\cite{cho2014learning} as the backbone, complemented by a multilayer perceptron (MLP) serving as the classification head. The architecture resembles the ET-RNN model and CoLES encoder~\cite{babaev2022coles, babaev2019rnn}. 
The aggregated hidden states of the backbone are used as inputs for the classification head during the training phase and as embeddings in the validation tasks. As an objective function, we employ the cross-entropy loss. 

\subsection{Contrastive models for transaction sequences}
\label{sec:contrastive_methods}

\paragraph{\textbf{CoLES method}}\label{sec:coles}
As a starting point in our study of SSL methods, we chose CoLES~\cite{babaev2022coles}, a representation learning method for discrete event sequences. 
Below is a brief overview of this approach; see the original paper for more details.

As all contrastive SSL approaches~\cite{liu2023ssl}, the CoLES model requires sets of positive and negative pairs for training. 
The authors of the method studied various ways to obtain such pairs. Eventually, they opted for the split strategy that considers two transaction subsequences from the same client to be a positive pair and from different clients --- to be a negative one.
This strategy does not disrupt the transactions' order, preserving their temporal structure. Representations of the subsequences were obtained via a Long Short-Term Memory network~\cite{hochreiter1997long} (LSTM) as an encoder model, which was trained using a contrastive loss~\cite{hadsell2006contrastiveloss}. In this work, we follow the same training pipeline as the original article.

\paragraph{\textbf{TS2Vec model}}\label{sec:ts2vec}
The contrastive model TS2Vec~\cite{yue2022ts2vec} has emerged as one of the leading approaches for obtaining representations of classical multivariate time series data. 
TS2Vec stands out for its distinct positive and negative sample selection strategy and hierarchical aggregation to facilitate representation learning at different scales. 

The authors have introduced contextual consistency, a novel approach for generating positive pairs. This approach designates representations at the same timestamp within two augmented contexts as positive pairs. 
Concerning negative samples, selection occurs in both instance-wise and temporal dimensions, complementing each other. 

In the initial step, the described contrastive approach is employed on timestamp-level representations. Following this, max pooling is executed along the temporal axis. Subsequently, the contrastive loss is applied to the acquired representations, and this entire process iteratively continues for each level until instance-level representations are attained. This mechanism proficiently empowers the model to systematically capture contextual information at diverse resolutions within temporal data. In our study, we rely on the original implementation of this method (with several modifications for transactional data type).

\paragraph{\textbf{Limitations of the considered contrastive approaches}}
The methods outlined above are only partially suited for modeling event sequences and thus have limitations.

The split strategy employed in CoLES explicitly encourages the model to build representations describing the user as a whole rather than capturing his local state at a specific moment. 
Moreover, the original paper only evaluated the global properties of the obtained embeddings since the users' classification was the primary downstream task.

TS2Vec, in turn, was primarily designed for standard time series samples with a uniform step. 
Therefore, it might be unfit for transactional data featuring irregular time intervals between observations~\cite{chowdhury2023primenet}. 
In our work, we explore the adaptability of TS2Vec to transaction history representation learning and incorporate it as a baseline model for comparative analysis.

\subsection{Our approaches to generative modeling for transaction sequences}\label{sec:ae}

\paragraph{\textbf{Classic autoencoder model}}

\begin{figure}[!ht]
    \centering
    \includegraphics[width=\textwidth]{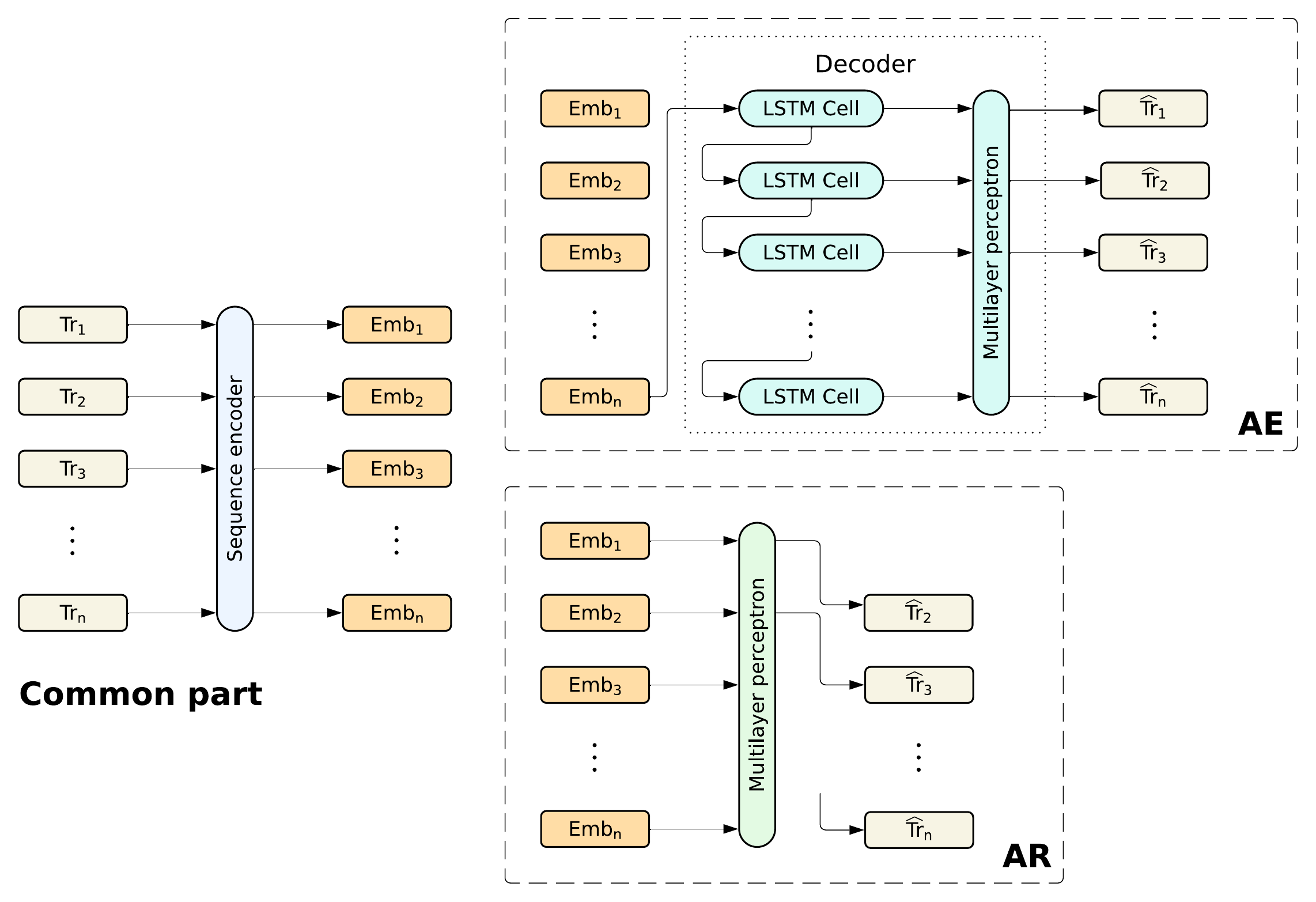}
    \caption{Scemes of the Autoencoder (AE) and Autoregressive (AR) methods.}
    \label{fig:ae_nlp_method}
\end{figure}

The autoencoder (AE)~\cite{tschannen2018recent} is a popular choice for generative models. 
Following this approach, the sequence is first encoded into a low-dimensional representation and then restored from it. 
The overall process is depicted in Figure~\ref{fig:ae_nlp_method}. 
Both the encoder and the decoder networks are represented by LSTMs~\cite{hochreiter1997long}, so the backbone model architecture is identical to CoLES. We add two linear heads to the decoder for amounts and MCC prediction.

Since transactional records contain both categorical (MCCs) and continuous (amounts) information, the reconstruction loss is divided into two parts. 
We use cross-entropy for the categorical features and mean squared error for the continuous features. 
The final loss function is a weighted sum of the two intermediate ones, with the weights being hyperparameters.

We found that the preprocessing of transaction amounts defined below is essential for the model to train successfully:
\begin{equation} \label{eq:ae_amount_transform}
    f(a) = \mathrm{sign}(a) \ln{(1 + |a|)}.
\end{equation}
On the one hand, this transformation allows us to balance out the loss functions for amounts and MCCs without assigning them drastically different weights. On the other, it better reflects the nature of human perception: we tend to focus on the order of magnitude, ignoring the precise value. 

For self-supervised approaches, it is very important to select the optimal complexity for the training objective~\cite{kenton2019bert, he2022masked}. 
In our case, it turns out that the prediction of rare MCC codes is an unbearable problem. 
So, to simplify the training objective, the number of unique MCC codes is reduced to 100 for all datasets by clipping all less frequent MCC codes. 

\paragraph{\textbf{Masked language model}}
\label{sec:mlm}
Diving deeper into generative methods, we also explore the masked language model (MLM), closely related to transformers~\cite{kenton2019bert, padhi2021tabular,vaswani2017attention}. The main components of transformers are self-attention and feed-forward blocks.

\begin{figure}[!ht]
    \centering
    \includegraphics[width=\textwidth]{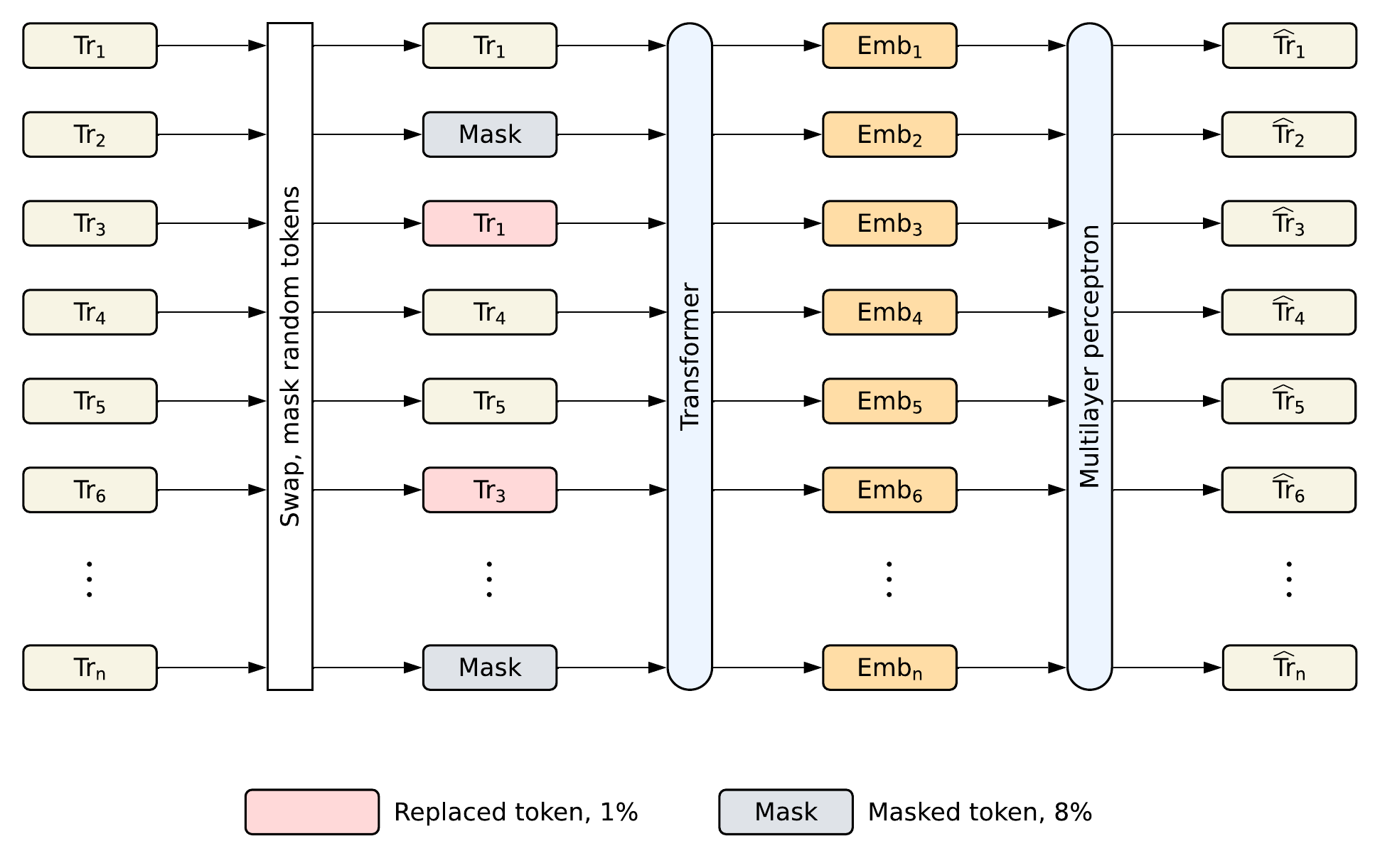}
    \caption{Scheme of the Masked Language Model (MLM) method.}
    \label{fig:mlm_method}
\end{figure}

An MLM works by reconstructing randomly masked tokens. The method is depicted in~\autoref{fig:mlm_method}. Before feeding the sequence to our model, we randomly select 10\% of the incoming tokens and preprocess them similarly to~\cite{kenton2019bert}. 
In particular, out of these 10\%, we:
\begin{itemize}
    \item mask 80\%, swapping out the MCC code for a special token and zeroing out the transaction amount;
    \item swap 10\% for another random transaction;
    \item and keep as is the remaining 10\%.
\end{itemize}

Like in the AE model, we use multiple linear layers to acquire the predictions from the encoder embeddings. 
The loss function is also a combination of cross-entropy and mean-squared error, but here, it is calculated only for the 10\% subset we sampled above. 
We also clip our MCC codes to 100 ones and preprocess the amounts like for the AE model.

It should also be noted that all transformer embeddings have equal receptive fields, each covering the whole sequence (which is not correct for RNN embeddings we use in other methods).
This implies that, in essence, any of these representations can be used in the downstream and validation tasks as representations of all the given tokens.
Following previous work on transformers, we choose the first token's embedding since that is traditionally where the \textit{CLS} token is.

\paragraph{\textbf{Autoregressive approach}}\label{sec:ar}
Autoregressive (AR) modeling is extensively used in CV~\cite{chen2018pixelsnail,esser2021taming, razavi2019generating, van2016pixel}, as well as in sequential domains like NLP~\cite{black2022gpt, brown2020language, radford2018improving} and Audio~\cite{borsos2023audiolm, oord2016wavenet}. 
Such models aim to predict the next item in a sequence. 
In contrast to token embeddings in autoencoders, those in autoregressive models do not use information regarding future tokens. 
This characteristic empowers autoregressive models to capture more intricate patterns, as demonstrated by their superior performance in text generation tasks~\cite{ethayarajh2019contextual}.

In our experiments, we employ a recurrent neural network similar to the CoLES model. We utilize the model's last hidden state to represent the transaction sequence since it encapsulates complete information about the entire sequence.
The model is trained to predict the next transaction, encompassing the MCC code and transaction amount, as illustrated in Figure~\ref{fig:ae_nlp_method}. 
In terms of transaction preprocessing and the loss function, they resemble the techniques employed in autoencoders, outlined in Section~\ref{sec:ae}. 

\subsection{Temporal Point Processing models}
\label{sec:methods_tpp}
When working with irregular time series or event sequences, it is crucial to consider that the records (measurements, indicators) have distinct time steps between each other. Such a problem statement is standard for TPP modeling. 

In this work, we choose three state-of-the-art models as baselines of that type: the Neural Hawkes Process (NHP)~\cite{mei2017neural}, the Attentive Neural Hawkes Process (A-NHP)~\cite{mei2021transformer}, and the COTIC~\cite{zhuzhel2023continuous} model. 
All these approaches follow a similar logic: they utilize deep neural networks for conditional event intensity parametrization.
The main optimization objective here is the log-likelihood of the observed event sequences.
The models' architectures and peculiarities are briefly described below.

\paragraph{\textbf{Neural Hawkes and Attentive Neural Hawkes Process}}
\label{subsec:nhp_anhp}
The NHP~\cite{mei2017neural} model originates from the Hawkes Process~\cite{hawkes1971spectra} --- one of the classic ways of event sequence modeling. 
Its key concept is self-excitation, i.e., past events temporarily raise the probability of future events.
This impact is assumed to be positive, additive, and exponentially decaying over time.
NHP generalizes the assumption by determining the conditional event intensities from a hidden state of a recurrent neural network.
The authors propose a novel continuous-time LSTM~\cite{hochreiter1997long} architecture that allows querying event representations at any time, even if no actual event occurred at that moment.

The A-NHP approach~\cite{mei2021transformer} develops the same ideas and suggests replacing the original recurrent encoder with a simpler parallelizable attention-based architecture~\cite{vaswani2017attention}.
The paper proves such a solution is more efficient while it provides comparable or better validation results.

\paragraph{\textbf{COTIC model}}
\label{subsec:cotic}
Continuous 1D convolutional neural networks (CCNN) have also demonstrated their efficiency for TPP modeling~\cite{li2020learning, shi2021continuous}.
A recently proposed COTIC approach~\cite{zhuzhel2023continuous} utilizes a deep CCNN to obtain event sequence representations and parameterize the intensity function of the underlying process.
On top of it, COTIC applies two multi-layer perceptron heads for return time and event type prediction.
Notably, these heads are trained jointly with the core part of the model using task-specific loss functions.
This strategy may enforce the desired local properties of the embeddings obtained from the encoder. 
However, using only an encoder from the COTIC, without additional heads, might lead to a degradation in quality.
We refer interested readers to the paper itself~\cite{zhuzhel2023continuous} for more details on the architecture and the losses.

\subsection{Usage of external information based on local representations of transactional data}\label{sec:global_context_methods}
\begin{figure}[!ht]
     \centering
     \includegraphics[width=0.75\columnwidth]{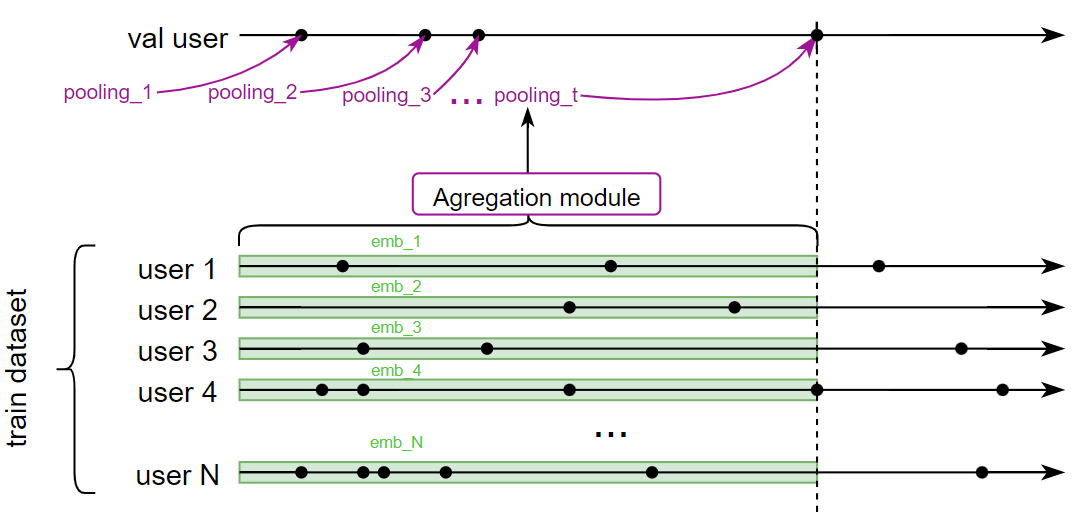}
     \caption{General pipeline for obtaining external context vectors.}
     \label{fig:global_pooling}
\end{figure}

According to the current research~\cite{ala2022deep, babaev2022coles}, it is beneficial to consider the macro context around users. 
We propose to construct an external context representation vector by properly aggregating the embeddings obtained from all or some selected users. 
The procedure is presented in Figure~\ref{fig:global_pooling} and is described as follows:

\begin{enumerate}
     \item We build local representations for all users and each unique transaction moment.
     \item For each client, we select all local embeddings from the dataset close to the current time point but before it.
     \item Obtained vectors are aggregated in the resulting vector of the external context representation.
\end{enumerate}

To validate the ability to improve the representations via external information, we concatenate each user’s local embeddings with this resulting vector and follow our general validation pipeline (see Section~\ref{chap:method} for details). 

In the experiments with the external context aggregation, we use CoLES as an encoder for building local representations. 
However, potentially, any model that transforms a sequence of transactions into a sequence of their representations fits our pipeline.

\paragraph{\textbf{Classical methods of context vector aggregation}}
Averaging and maximization were used as classical aggregation methods. Our motivation comes from an analogy with applying Mean and Max Pooling operations in convolutional neural networks~\cite{boureau2010theoretical}.
In the first case, we obtain the external context vector by component-wise averaging the local vectors for all users in the stored dataset, and in the second case, we take the maximum value for each component. Like in convolutional networks, such aggregation methods are designed to generalize the environment for the current point.

\paragraph{\textbf{Methods based on the attention mechanism}}
Following a fruitful idea from the Recommender Systems domain, we suppose that some bank clients may behave more similarly than others. 
Thus, such people determine the user's closest environment and help to describe his behavior better. 
The aggregation methods described above do not consider the users' representation relationships and, as a consequence, their similarity. 
To overcome this drawback, we propose applying the attention mechanism. 
It was initially designed to account for interword similarities during context aggregation in the machine translation task ~\cite{vaswani2017attention} that resembles our problem. 

In this work, we utilize two variants of the attention mechanism: without a learnable attention matrix and with it. 
The corresponding form for the external context vectors for a given point in time are:
\begin{equation}
     \boldsymbol{B}_t = X\sigma(X^T \mathbf{h}_t),\quad
     \boldsymbol{B}_t = X\sigma(X^T A \mathbf{h}_t),
\end{equation}
where $\mathbf{h}_t \in \mathbb{R}^{m}$ is a vector of local representation for the considered client; $X \in \mathbb{R}^{m \times n}$ is a matrix, which columns are embeddings of all $n$ users (except considered) at a given time; $A \in \mathbb{R}^{m \times m}$ is a trainable matrix; and $\sigma$ denotes the softmax function $\sigma(\mathbf{z})_i = \frac{\exp(z_i)}{\sum_{j = 1}^d \exp{(z_j)}}$. 

In the first case, the user similarity metric is the softmax of the scalar product. 
In the second, vectors of representations from the dataset are additionally passed through a trained linear layer. 
We train the attention matrix via the CoLES model training pipeline (see Section~\ref{sec:contrastive_methods} for more information).

%% file: 5-data.tex
In this study, we compare our methods using public and private datasets of transaction records from different banks' customers. 
We briefly describe the data below and provide the main technical characteristics of open datasets in Table \ref{tab:datasets}.

\paragraph{\textbf{Churn}}
The  \href{https://boosters.pro/championship/rosbank1}{Churn} dataset was initially dedicated to identifying potential customers about to leave. We also use this problem for our global validation (Section~\ref{chap:method}). Due to the nature of such events, the target value distribution is unbalanced. For the local validation tasks, we assume that clients change their behavior one month before their last transaction.

\paragraph{\textbf{Default}}
\href{https://boosters.pro/championship/alfabattle2}{Default} has a resembling structure to the Churn problem. The corresponding global downstream task is to classify clients as able to repay the bank loan. This task is close to the real-world business problem of credit scoring. Likewise, we mark transactions before the default month as "pre-default" for local validation. It is also important to note that this dataset has a significant class imbalance, which may lead to increased standard deviations of the metrics.

\paragraph{\textbf{HSBC}}
The \href{https://www.kaggle.com/datasets/ashisparida/hsbc-ml-hackathon-2023}{HSBC} competition proposes the fraud identification task.
The original dataset contains local targets indicating if each transaction was made by the client personally or if it was malicious. 
We also add client-level global tags showing that at least one operation in a transaction history sequence was fake. 
Obviously, there are far fewer labels indicating that transactions are fraudulent, so the target variable is highly imbalanced.

\paragraph{\textbf{Age}}
The Age \href{https://ods.ai/competitions/sberbank-sirius-lesson}{dataset} implies predicting bank clients' age group based on their transactions history. 
Four groups are in total, creating a relatively balanced multi-class global downstream task. 
The Age differs from other considered datasets. 
Its samples do not include exact timestamps for each transaction. 
Instead, they only have a serial number of the day the transaction is made starting from some base date. 
In addition, there is no natural local task for this data (since the global target's dynamic is negligible in the considered period), so we skip this local downstream validation step. 

\paragraph{\textbf{Real-world private dataset}}
A major bank has provided us with a proprietary dataset containing the anonymized data of a subset of real clients. The total number of unique customers in this data is five million, with a time range of approximately one year. 
This dataset includes characteristics such as each client's age and gender, which we use for both binary and multi-class classification during the global validation process. Similarly to Age, we skip local downstream task validation and evaluate only performance on the MCC prediction. 

\begin{table}[!ht]
\caption{Basic statistics of the open datasets used in the experiments.}
\label{tab:datasets}
\centering
\begin{tabular}{lccccc} \hline
~ & Churn & Default & HSBC & Age \\ \hline
\# of transactions & 490K & 2M & 234K & 26M \\ 
\# of clients & 5K & 7K & 4K & 30K \\
Min transaction length & 1 & 300 & 1 & 700 \\
Max transaction length & 784 & 300 & 3467 & 1150 \\
Median transaction length & 83 & 300 & 40 & 863 \\
\# of MCCs & 344 & 309 & 307 & 202 \\
\# of global classes & 2 & 2 & 2 & 4  \\
Class balance & yes & no & no & yes \\ \hline
\end{tabular}
\end{table}
\paragraph{\textbf{Change point detection datasets}}

CPD experiments are conducted on both real and synthetic data. 
For the real-world scenario, we use a subset of $35$ clients from the Churn dataset containing long transaction sequences with a currency change serving as the change point. Since the real data with annotated change points is limited, we also consider synthetic sequences. For this purpose, we constructed samples by combining subsequences from different clients.

\paragraph{\textbf{Data preprocessing}}
\label{subsec:data_preprocessing}
We carry out minimal preprocessing for analyzed datasets. Our steps include bringing all timestamps to one format, removing unnecessary columns, and encoding MCCs by frequency. 
Based on the above logic (Subsection~\ref{subsec:local_validation}), we also add local binary targets for each of the three datasets: Churn, Default, and HSBC.

To correctly evaluate our models' performance, we split the data into three parts: training set (80\% of all users), validation set (10\%), and test set (10\%). In our pipeline, the encoder models use only the training set for learning. Regarding validation, additional local classifiers (MLP heads) use the same training/validation/test split as the encoders. In turn, global validation classifiers (gradient boosting) are trained on training and validation sets and evaluated on the test set. 

%% file: 6-results.tex
\label{chap:experiments}
Below, we provide the results we achieved during our work.
The implications of each figure and table are discussed meticulously in the corresponding text.
This section concludes with an executive summary, which contains a ranking of the considered models, as well as a brief analysis of said ranking.

Within the framework of our project, we compare several classes of transactional data models described in Section~\ref{chap:models}.
Contrastive self-supervised approaches include CoLES~\cite{babaev2022coles} and TS2Vec~\cite{yue2022ts2vec}; 
generative self-supervised methods are presented by the classic autoencoder (AE) (Subsection~\ref{sec:ae}), the masked language model (MLM) (Subsection~\ref{sec:mlm}), and the autoregressive model (AR) (Subsection~\ref{sec:ar});
and Temporal Point Process baselines encompass the NHP~\cite{mei2017neural}, the A-NHP~\cite{mei2021transformer}, and the COTIC~\cite{zhuzhel2023continuous} models.
Additionally, we include the results for the best supervised baseline (Best sup.), which was trained using the global markup. 
It is important to note that these models could be superior to the self-supervised ones in many cases, as they directly use the correct labels.

The details on the models' architecture, implementation, and hyperparameters are given in ~\ref{sec:implementation_details}.
We release the GitHub repository\footnote{\url{https://github.com/romanenkova95/transactions_gen_models}} containing the reproducible Python code for all the experiments conducted in this research.

\subsection{Analysis of local and global properties of representations}
\label{sec:global_local_quality}

\begin{figure}[!ht]
     \centering
     \includegraphics[width=\textwidth]{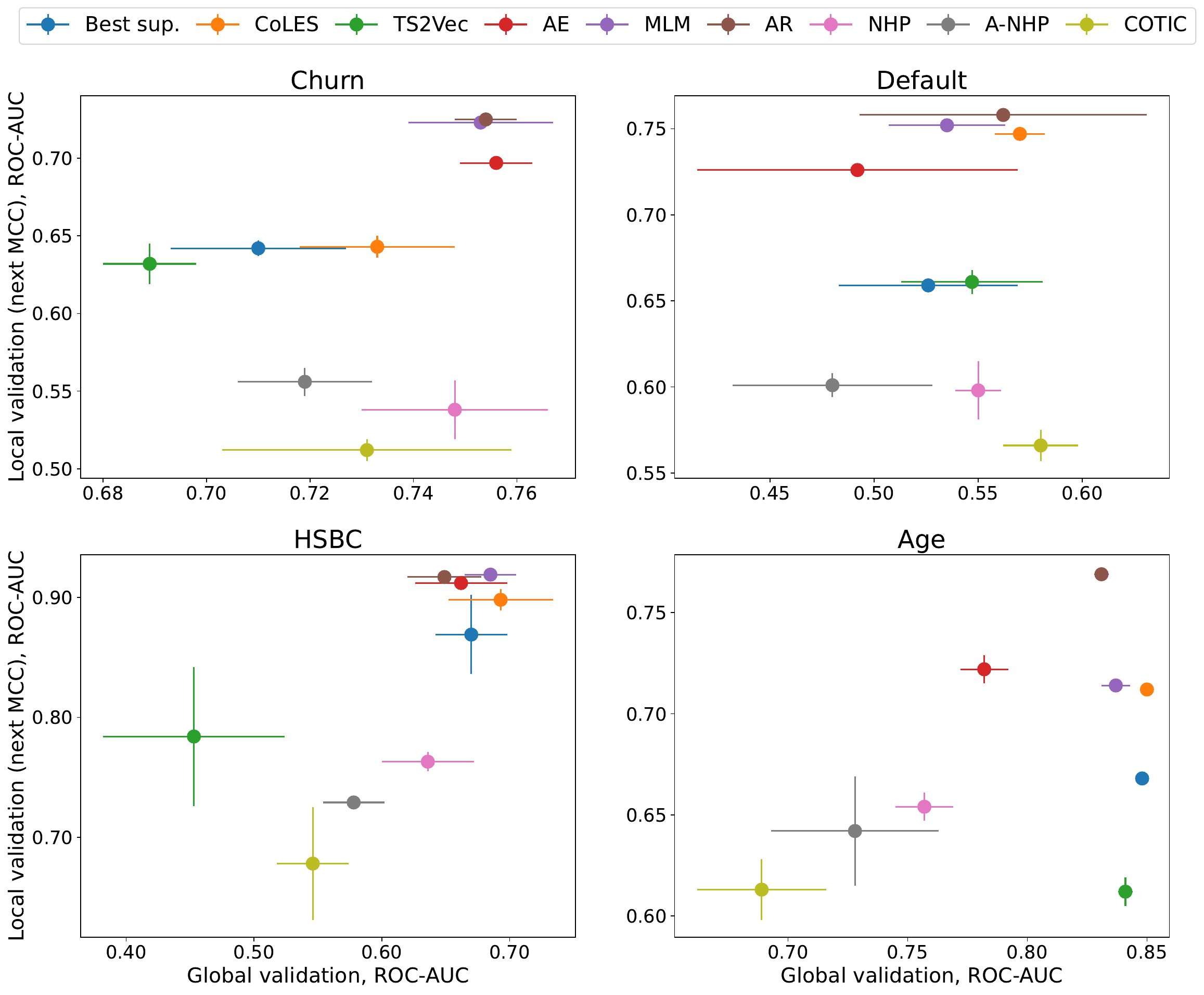}
     \caption{Quality of the models regarding their global and local properties. ``Best sup.'' stands for the best supervised baseline. The x-axis at each graph corresponds to global validation ROC-AUC, while the y-axis shows the next MCC prediction ROC-AUC. Thus, the upper and righter the dot is, the better model it represents.}
     \label{fig:local_vs_global_quality}
\end{figure}

The main results for the four open datasets, emphasizing the trade-off between local and global properties of the models, are illustrated in Figure~\ref{fig:local_vs_global_quality}. 
More details are provided in Tables~\ref{tab:main_results}.

In these graphs, the x-axis represents the global benchmark, while the y-axis represents the local one (predicting the next MCC code).
Thus, the closer the dot is to the upper right corner, the better the corresponding model.

Note that the two datasets under consideration (Default and HSBC) are highly imbalanced.
Consequently, the accuracy is not a relevant classification metric in this case. 
Thus, we suggest that the integral metrics (ROC-AUC and PR-AUC) are the main quality measures in the experiments.

\begin{table}[!ht]
\begin{adjustbox}{width=\textwidth}
\begin{tabular}{l|ccc|ccc|ccc}
\hline
\multirow{2}{*}{Method} & 
\multicolumn{3}{c|}{Global validation} & 
\multicolumn{3}{c|}{Local validation: next MCC} & 
\multicolumn{3}{c}{Local validation: binary target} \\ \cline{2-10}
& Accuracy & ROC-AUC & PR-AUC & Accuracy & ROC-AUC & PR-AUC & Accuracy & ROC-AUC & PR-AUC \\ \hline
\multicolumn{10}{c}{Churn} \\ \hline
Best sup. & 0.67\scriptsize{$\pm$0.02} & 0.71\scriptsize{$\pm$0.02} & 0.75\scriptsize{$\pm$0.02} & 0.25\scriptsize{$\pm$0.01} & 0.64\scriptsize{$\pm$0.01} & 0.17\scriptsize{$\pm$0.00} & 0.73\scriptsize{$\pm$0.01} & 0.55\scriptsize{$\pm$0.01} & 0.31\scriptsize{$\pm$0.01} \\ \hdashline
CoLES & 0.67\scriptsize{$\pm$0.02} & 0.73\scriptsize{$\pm$0.02} & 0.77\scriptsize{$\pm$0.03} & 0.23\scriptsize{$\pm$0.00} & 0.64\scriptsize{$\pm$0.01} & 0.16\scriptsize{$\pm$0.01} & \textbf{0.73}\scriptsize{$\pm$0.00} & 0.56\scriptsize{$\pm$0.01} & 0.32\scriptsize{$\pm$0.01} \\
TS2Vec & 0.65\scriptsize{$\pm$0.02} & 0.69\scriptsize{$\pm$0.01} & 0.75\scriptsize{$\pm$0.02} & 0.16\scriptsize{$\pm$0.01} & 0.63\scriptsize{$\pm$0.01} & 0.14\scriptsize{$\pm$0.01} & \underline{0.65}\scriptsize{$\pm$0.05} & 0.52\scriptsize{$\pm$0.03} & 0.28\scriptsize{$\pm$0.02} \\
AE & \underline{0.69}\scriptsize{$\pm$0.01} & \textbf{0.76}\scriptsize{$\pm$0.01} & \textbf{0.82}\scriptsize{$\pm$0.01} & \underline{0.27}\scriptsize{$\pm$0.01} & 0.70\scriptsize{$\pm$0.00} & \underline{0.21}\scriptsize{$\pm$0.00} & \textbf{0.73}\scriptsize{$\pm$0.00} & \underline{0.57}\scriptsize{$\pm$0.00} & 0.33\scriptsize{$\pm$0.02} \\
MLM & 0.67\scriptsize{$\pm$0.01} & \underline{0.75}\scriptsize{$\pm$0.01} & \underline{0.81}\scriptsize{$\pm$0.02} & \textbf{0.28}\scriptsize{$\pm$0.00} & \underline{0.72}\scriptsize{$\pm$0.00} & \underline{0.21}\scriptsize{$\pm$0.01} & \textbf{0.73}\scriptsize{$\pm$0.00} & \underline{0.57}\scriptsize{$\pm$0.01} & 0.33\scriptsize{$\pm$0.01} \\
AR & \textbf{0.70}\scriptsize{$\pm$0.00} & \underline{0.75}\scriptsize{$\pm$0.01} & 0.77\scriptsize{$\pm$0.01} & \textbf{0.28}\scriptsize{$\pm$0.00} & \textbf{0.73}\scriptsize{$\pm$0.00} & \textbf{0.23}\scriptsize{$\pm$0.00} & \textbf{0.73}\scriptsize{$\pm$0.00} & \textbf{0.58}\scriptsize{$\pm$0.01} & 0.34\scriptsize{$\pm$0.01} \\
NHP & 0.68\scriptsize{$\pm$0.02} & \underline{0.75}\scriptsize{$\pm$0.02} & 0.77\scriptsize{$\pm$0.01} & 0.24\scriptsize{$\pm$0.00} & 0.54\scriptsize{$\pm$0.02} & 0.13\scriptsize{$\pm$0.01} & 0.63\scriptsize{$\pm$0.00} & 0.54\scriptsize{$\pm$0.04} & \underline{0.40}\scriptsize{$\pm$0.04} \\
A-NHP & 0.67\scriptsize{$\pm$0.02} & 0.72\scriptsize{$\pm$0.01} & 0.75\scriptsize{$\pm$0.01} & 0.21\scriptsize{$\pm$0.01} & 0.56\scriptsize{$\pm$0.01} & 0.13\scriptsize{$\pm$0.01} & 0.64\scriptsize{$\pm$0.00} & 0.55\scriptsize{$\pm$0.03} & \textbf{0.41}\scriptsize{$\pm$0.02} \\
COTIC & 0.67\scriptsize{$\pm$0.03} & 0.73\scriptsize{$\pm$0.03} & 0.75\scriptsize{$\pm$0.02} & 0.22\scriptsize{$\pm$0.03} & 0.51\scriptsize{$\pm$0.01} & 0.12\scriptsize{$\pm$0.00} & 0.63\scriptsize{$\pm$0.00} & 0.49\scriptsize{$\pm$0.01} & 0.36\scriptsize{$\pm$0.01} \\ \hline

\multicolumn{10}{c}{Default} \\ \hline
Best sup. & 0.91\scriptsize{$\pm$0.00} & 0.53\scriptsize{$\pm$0.04} & 0.09\scriptsize{$\pm$0.03} & 0.31\scriptsize{$\pm$0.00} & 0.66\scriptsize{$\pm$0.00} & 0.21\scriptsize{$\pm$0.01} & 0.90\scriptsize{$\pm$0.01} & 0.50\scriptsize{$\pm$0.01} & 0.01\scriptsize{$\pm$0.00} \\ \hdashline
CoLES & 0.86\scriptsize{$\pm$0.02} & \underline{0.57}\scriptsize{$\pm$0.01} & 0.06\scriptsize{$\pm$0.01} & \underline{0.34}\scriptsize{$\pm$0.01} & \underline{0.75}\scriptsize{$\pm$0.00} & \underline{0.26}\scriptsize{$\pm$0.00} & 0.93\scriptsize{$\pm$0.08} & 0.54\scriptsize{$\pm$0.04} & 0.01\scriptsize{$\pm$0.00} \\
TS2Vec & \textbf{0.93}\scriptsize{$\pm$0.00} & 0.55\scriptsize{$\pm$0.03} & \textbf{0.09}\scriptsize{$\pm$0.02} & 0.25\scriptsize{$\pm$0.02} & 0.66\scriptsize{$\pm$0.01} & 0.18\scriptsize{$\pm$0.01} & \textbf{0.99}\scriptsize{$\pm$0.00} & \textbf{0.61}\scriptsize{$\pm$0.05} & 0.01\scriptsize{$\pm$0.00} \\
AE & 0.56\scriptsize{$\pm$0.02} & 0.49\scriptsize{$\pm$0.08} & 0.05\scriptsize{$\pm$0.01} & 0.33\scriptsize{$\pm$0.01} & 0.73\scriptsize{$\pm$0.00} & \underline{0.26}\scriptsize{$\pm$0.00} & 0.96\scriptsize{$\pm$0.00} & 0.53\scriptsize{$\pm$0.05} & 0.01\scriptsize{$\pm$0.00} \\
MLM & 0.73\scriptsize{$\pm$0.02} & 0.54\scriptsize{$\pm$0.03} & 0.06\scriptsize{$\pm$0.01} & \underline{0.34}\scriptsize{$\pm$0.00} & \underline{0.75}\scriptsize{$\pm$0.00} & \underline{0.26}\scriptsize{$\pm$0.00} & \textbf{0.99}\scriptsize{$\pm$0.00} & 0.56\scriptsize{$\pm$0.04} & 0.01\scriptsize{$\pm$0.00} \\
AR & 0.87\scriptsize{$\pm$0.01} & 0.56\scriptsize{$\pm$0.07} & \underline{0.08}\scriptsize{$\pm$0.02} & \textbf{0.35}\scriptsize{$\pm$0.00} & \textbf{0.76}\scriptsize{$\pm$0.00} & \textbf{0.28}\scriptsize{$\pm$0.00} & 0.97\scriptsize{$\pm$0.02} & 0.44\scriptsize{$\pm$0.05} & 0.01\scriptsize{$\pm$0.00} \\
NHP & \underline{0.92}\scriptsize{$\pm$0.01} & 0.55\scriptsize{$\pm$0.01} & 0.07\scriptsize{$\pm$0.02} & 0.30\scriptsize{$\pm$0.00} & 0.60\scriptsize{$\pm$0.02} & 0.15\scriptsize{$\pm$0.01} & \textbf{0.99}\scriptsize{$\pm$0.00} & 0.52\scriptsize{$\pm$0.04} & 0.01\scriptsize{$\pm$0.00} \\
A-NHP & 0.72\scriptsize{$\pm$0.02} & 0.48\scriptsize{$\pm$0.05} & 0.04\scriptsize{$\pm$0.01} & 0.29\scriptsize{$\pm$0.01} & 0.60\scriptsize{$\pm$0.01} & 0.16\scriptsize{$\pm$0.00} & \underline{0.98}\scriptsize{$\pm$0.01} & \underline{0.60}\scriptsize{$\pm$0.06} & 0.01\scriptsize{$\pm$0.00} \\
COTIC & 0.90\scriptsize{$\pm$0.01} & \textbf{0.58}\scriptsize{$\pm$0.02} & \underline{0.08}\scriptsize{$\pm$0.02} & 0.30\scriptsize{$\pm$0.00} & 0.57\scriptsize{$\pm$0.01} & 0.14\scriptsize{$\pm$0.01} & 0.94\scriptsize{$\pm$0.07} & 0.38\scriptsize{$\pm$0.02} & 0.00\scriptsize{$\pm$0.00} \\ \hline

\multicolumn{10}{c}{HSBC} \\ \hline
Best sup. & 0.92\scriptsize{$\pm$0.00} & 0.67\scriptsize{$\pm$0.03} & 0.15\scriptsize{$\pm$0.01} & 0.71\scriptsize{$\pm$0.01} & 0.87\scriptsize{$\pm$0.03} & 0.78\scriptsize{$\pm$0.04} & 1.00\scriptsize{$\pm$0.00} & 0.37\scriptsize{$\pm$0.13} & 0.01\scriptsize{$\pm$0.00} \\ \hdashline
CoLES & 0.92\scriptsize{$\pm$0.00} & \textbf{0.69}\scriptsize{$\pm$0.04} & \underline{0.15}\scriptsize{$\pm$0.02} & 0.69\scriptsize{$\pm$0.02} & 0.90\scriptsize{$\pm$0.01} & 0.82\scriptsize{$\pm$0.01} & \textbf{1.00}\scriptsize{$\pm$0.00} & 0.38\scriptsize{$\pm$0.09} & 0.01\scriptsize{$\pm$0.00} \\
TS2Vec & \textbf{0.95}\scriptsize{$\pm$0.04} & 0.45\scriptsize{$\pm$0.07} & 0.05\scriptsize{$\pm$0.04} & 0.65\scriptsize{$\pm$0.04} & 0.78\scriptsize{$\pm$0.06} & 0.66\scriptsize{$\pm$0.04} & \underline{0.97}\scriptsize{$\pm$0.04} & 0.54\scriptsize{$\pm$0.05} & \textbf{0.05}\scriptsize{$\pm$0.07} \\
AE & 0.92\scriptsize{$\pm$0.00} & 0.66\scriptsize{$\pm$0.04} & 0.14\scriptsize{$\pm$0.03} & 0.72\scriptsize{$\pm$0.04} & \underline{0.91}\scriptsize{$\pm$0.01} & \underline{0.84}\scriptsize{$\pm$0.02} & \textbf{1.00}\scriptsize{$\pm$0.00} & 0.44\scriptsize{$\pm$0.09} & 0.01\scriptsize{$\pm$0.00} \\
MLM & \underline{0.93}\scriptsize{$\pm$0.00} & \textbf{0.69}\scriptsize{$\pm$0.02} & \underline{0.15}\scriptsize{$\pm$0.02} & \textbf{0.80}\scriptsize{$\pm$0.01} & \textbf{0.92}\scriptsize{$\pm$0.00} & \textbf{0.85}\scriptsize{$\pm$0.00} & \textbf{1.00}\scriptsize{$\pm$0.00} & \textbf{0.81}\scriptsize{$\pm$0.02} & \underline{0.02}\scriptsize{$\pm$0.00} \\
AR & 0.92\scriptsize{$\pm$0.00} & 0.65\scriptsize{$\pm$0.03} & \underline{0.15}\scriptsize{$\pm$0.01} & \underline{0.79}\scriptsize{$\pm$0.02} & \textbf{0.92}\scriptsize{$\pm$0.00} & \textbf{0.85}\scriptsize{$\pm$0.00} & \textbf{1.00}\scriptsize{$\pm$0.00} & \underline{0.79}\scriptsize{$\pm$0.02}	& \underline{0.02}\scriptsize{$\pm$0.01} \\
NHP & 0.92\scriptsize{$\pm$0.00} & 0.64\scriptsize{$\pm$0.04} & \textbf{0.19}\scriptsize{$\pm$0.04} & 0.66\scriptsize{$\pm$0.00} & 0.76\scriptsize{$\pm$0.01} & 0.67\scriptsize{$\pm$0.02} & \textbf{1.00}\scriptsize{$\pm$0.00} & 0.48\scriptsize{$\pm$0.02} & 0.01\scriptsize{$\pm$0.01} \\
A-NHP & \underline{0.93}\scriptsize{$\pm$0.00} & 0.58\scriptsize{$\pm$0.02} & \underline{0.15}\scriptsize{$\pm$0.01} & 0.66\scriptsize{$\pm$0.00} & 0.73\scriptsize{$\pm$0.00} & 0.63\scriptsize{$\pm$0.00} & \textbf{1.00}\scriptsize{$\pm$0.00} & 0.32\scriptsize{$\pm$0.06} & 0.00\scriptsize{$\pm$0.00} \\
COTIC & 0.92\scriptsize{$\pm$0.00} & 0.55\scriptsize{$\pm$0.03} & 0.10\scriptsize{$\pm$0.02} & 0.65\scriptsize{$\pm$0.01} & 0.68\scriptsize{$\pm$0.05} & 0.64\scriptsize{$\pm$0.01} & 0.96\scriptsize{$\pm$0.02} & 0.51\scriptsize{$\pm$0.04} & 0.01\scriptsize{$\pm$0.00} \\ \hline

\multicolumn{10}{c}{Age} \\ \hline
Best sup. & 0.61\scriptsize{$\pm$0.00} & 0.85\scriptsize{$\pm$0.00} & 0.65\scriptsize{$\pm$0.00} & 0.33\scriptsize{$\pm$0.00} & 0.67\scriptsize{$\pm$0.00} & 0.22\scriptsize{$\pm$0.00} & ~ & ~ & ~ \\ \cdashline{1-7}
CoLES & \textbf{0.61}\scriptsize{$\pm$0.00} & \textbf{0.85}\scriptsize{$\pm$0.00} & \textbf{0.66}\scriptsize{$\pm$0.00} & 0.32\scriptsize{$\pm$0.00} & 0.71\scriptsize{$\pm$0.00} & 0.24\scriptsize{$\pm$0.00} & ~ & ~ & ~ \\
TS2Vec & \underline{0.60}\scriptsize{$\pm$0.01} & \underline{0.84}\scriptsize{$\pm$0.00} & \underline{0.64}\scriptsize{$\pm$0.00} & 0.18\scriptsize{$\pm$0.02} & 0.61\scriptsize{$\pm$0.01} & 0.17\scriptsize{$\pm$0.01} & ~ & ~ & ~ \\
AE & 0.52\scriptsize{$\pm$0.02} & 0.78\scriptsize{$\pm$0.01} & 0.54\scriptsize{$\pm$0.02} & \underline{0.35}\scriptsize{$\pm$0.00} & \underline{0.72}\scriptsize{$\pm$0.01} & \underline{0.26}\scriptsize{$\pm$0.00} &  ~ & ~ & ~ \\
MLM & 0.59\scriptsize{$\pm$0.01} & \underline{0.84}\scriptsize{$\pm$0.01} & 0.63\scriptsize{$\pm$0.01} & 0.33\scriptsize{$\pm$0.00} & 0.71\scriptsize{$\pm$0.00} & 0.24\scriptsize{$\pm$0.01} & ~ & NA & ~ \\
AR & 0.59\scriptsize{$\pm$0.01} & 0.83\scriptsize{$\pm$0.00} & 0.63\scriptsize{$\pm$0.01} & \textbf{0.37}\scriptsize{$\pm$0.00} & \textbf{0.77}\scriptsize{$\pm$0.00} & \textbf{0.30}\scriptsize{$\pm$0.00} & ~ & ~ & ~ \\
NHP & 0.49\scriptsize{$\pm$0.02} & 0.76\scriptsize{$\pm$0.01} & 0.52\scriptsize{$\pm$0.02} & 0.28\scriptsize{$\pm$0.00} & 0.65\scriptsize{$\pm$0.01} & 0.19\scriptsize{$\pm$0.01} &  ~ & ~ & ~ \\
A-NHP & 0.47\scriptsize{$\pm$0.03} & 0.73\scriptsize{$\pm$0.04} & 0.48\scriptsize{$\pm$0.04} & 0.28\scriptsize{$\pm$0.00} & 0.64\scriptsize{$\pm$0.03} & 0.18\scriptsize{$\pm$0.01} & ~ & ~ & ~ \\
COTIC & 0.43\scriptsize{$\pm$0.02} & 0.69\scriptsize{$\pm$0.03} & 0.43\scriptsize{$\pm$0.03} & 0.28\scriptsize{$\pm$0.00} & 0.61\scriptsize{$\pm$0.02} & 0.16\scriptsize{$\pm$0.01} & ~ & ~ & ~ \\ \hline
\end{tabular}
\end{adjustbox}
\caption{
Quality metrics for global and local embedding validation results. 
All metrics in the Table should be maximized. 
The results are averaged by three runs and are given in the format $mean \pm std$. 
The best values are \textbf{highlighted}, and the second best values are \underline{underlined}. 
``Best sup.'' stands for the best supervised baseline model. 
Note that there are no local binary targets for the Age dataset.
}
\label{tab:main_results}
\end{table}

The results (see Tables~\ref{tab:event_type_ranks},\ref{tab:local_binary_ranks},\ref{tab:global_ranks} for a summary) demonstrate that the generative models (AE, MLM, and AR) outperform the contrastive methods (CoLES and TS2Vec) at local validation tasks, aligning with our expectations and the generally accepted understanding of generative models~\cite{jaiswal2020survey}.
Importantly, they also typically keep up with the contrastive methods or even exceed the state-of-the-art CoLES approach on global tasks.

The more detailed conclusions are the following:
\begin{itemize}
    \item On the Churn dataset, the AE yields the best results among generative models for global tasks, but it is outperformed by AR in local tasks. 
    Contrastive approaches and TPP models are slightly behind in global quality but drastically underperform in the next MCC prediction.
    
    \item For Default, the outcomes are more nuanced: the best global results are for CoLES and COTIC, while AR leads among generative approaches. 
    For the next MCC prediction, the AR model is superior to others, followed by MLM.
    
    \item HSBC and Age results look similar to the Churn ones. However, CoLES and MLM are the best in terms of global properties. The local tasks are, again, dominated by the generative models (AR, MLM, and AE).
    
    \item In most cases, TPP models do not produce high-quality representations of event sequences. This issue can be explained as these methods generally aim at another task --- the intensity function restoration. 

    \item Surprisingly, supervised models do not provide efficient embeddings in most cases. This could be due to the noise in the markup of the datasets we used.
\end{itemize}

Overall, the generative models (AR and MLM, in particular) look preferable regarding the trade-off between the local and the global properties of their embeddings.
However, when selecting a specific solution, one must balance the models, considering the peculiarities of the downstream task at hand.

\paragraph{\textbf{Results on the proprietary data}}~\label{sec:proprietary}
In Table~\ref{tab:sber_results}, we provide the results of the experiments carried out on the industrial-scale private dataset described in Section~\ref{chap:data}.
Note that standard deviation is absent: due to the size of this dataset, it is infeasible to collect statistics over several runs.

\begin{table}[!ht]
\begin{adjustbox}{width=\textwidth}
\begin{tabular}{l|c|c|c} \hline
\multirow{2}{*}{Method} & \multirow{2}{*}{Global validation: age} & \multirow{2}{*}{Global validation: gender} & \multirow{2}{*}{Local validation: next MCC} \\ & & & \\ \hline
CoLES & \textbf{0.955} & \textbf{0.920} &\underline{0.750} \\
TS2Vec & 0.918 & 0.839 & 0.692 \\
AE & 0.886 & 0.753 & 0.734 \\
MLM & 0.939 & \underline{0.896} & 0.734 \\
AR & \underline{0.946} & 0.865 &  \textbf{0.770}\\
COTIC & 0.919 & 0.840 & 0.684 \\ \hline
\end{tabular}
\end{adjustbox}
\caption{
Global and local validation ROC-AUC values on the private industrial-scale dataset. All metrics in the Table should be maximized. The best values are \textbf{highlighted with bold font}, and the second best values are \underline{underlined}. 
}
\label{tab:sber_results}
\end{table}
The overall conclusion remains the same for this data: the generative model AR outperforms CoLES in the local task while falling slightly short of it on global properties. Another generative MLM approach also shows comparable results on both global and local problems.

\subsection{Experiments on the Dynamics of Representations}
\label{sec:dynamics_quality}

We adhere to our methodology for investigating the dynamic properties of representations introduced in Section~\ref{sec:cp_methods}. 
The experiments are conducted for three models: CoLES, AE, and AR. 

\paragraph{\textbf{Dynamics assessment using artificial data}}
\label{sec:dynamics_artificial_results}

The experiment is designed as follows. We consider $100$ pairs of clients. For each pair, we substitute the second half of one client's transactions with another's. As a result, the first client's history now contains an artificial change point. 
Then, we measure the distances between the clients' local representations: we assume they are significant before the change point and start to decrease right after it. 

We also perform the opposite experiment (by replacing the first half of the transactions instead of the second): initially, the synthesized clients' sequences are similar, but after the change point, their behavior diverges. The distances are expected to increase accordingly. 

\begin{figure}[!h]
    \centering
    \includegraphics[width=\columnwidth]{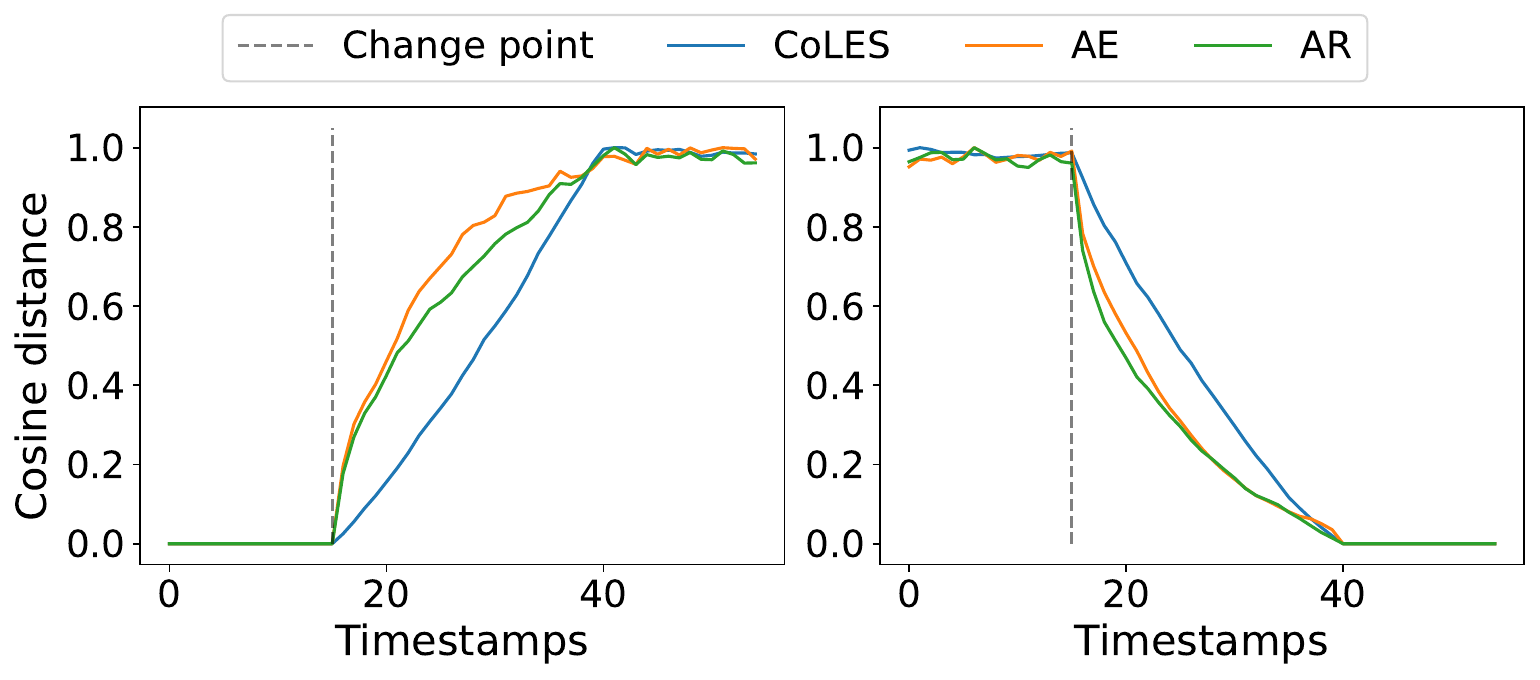}
    \caption{Distances between sequences' local representations. Left figure: clients' behavior diverges; right figure: it converges. Results are averaged over $100$ sequences.}
    \label{fig:sim_dif_change_point}
\end{figure}

The resulting dynamic for setups described above is depicted in Figure~\ref{fig:sim_dif_change_point}. Both AE and AR react to changes faster than CoLES. The figure also demonstrates a difference in convergence speed: for generative models, it appears to be exponentially fast, whereas for the contrastive one, it is closer to linear.

\paragraph{\textbf{Dynamics of real-world data embeddings}}

In this experiment, we use a small subset of the Churn dataset consisting of $35$ clients whose transaction history contains intrinsic change points. As such change moments, we consider changes in the transaction currency. 

\begin{figure}[!ht]
\includegraphics[width=0.85\textwidth]{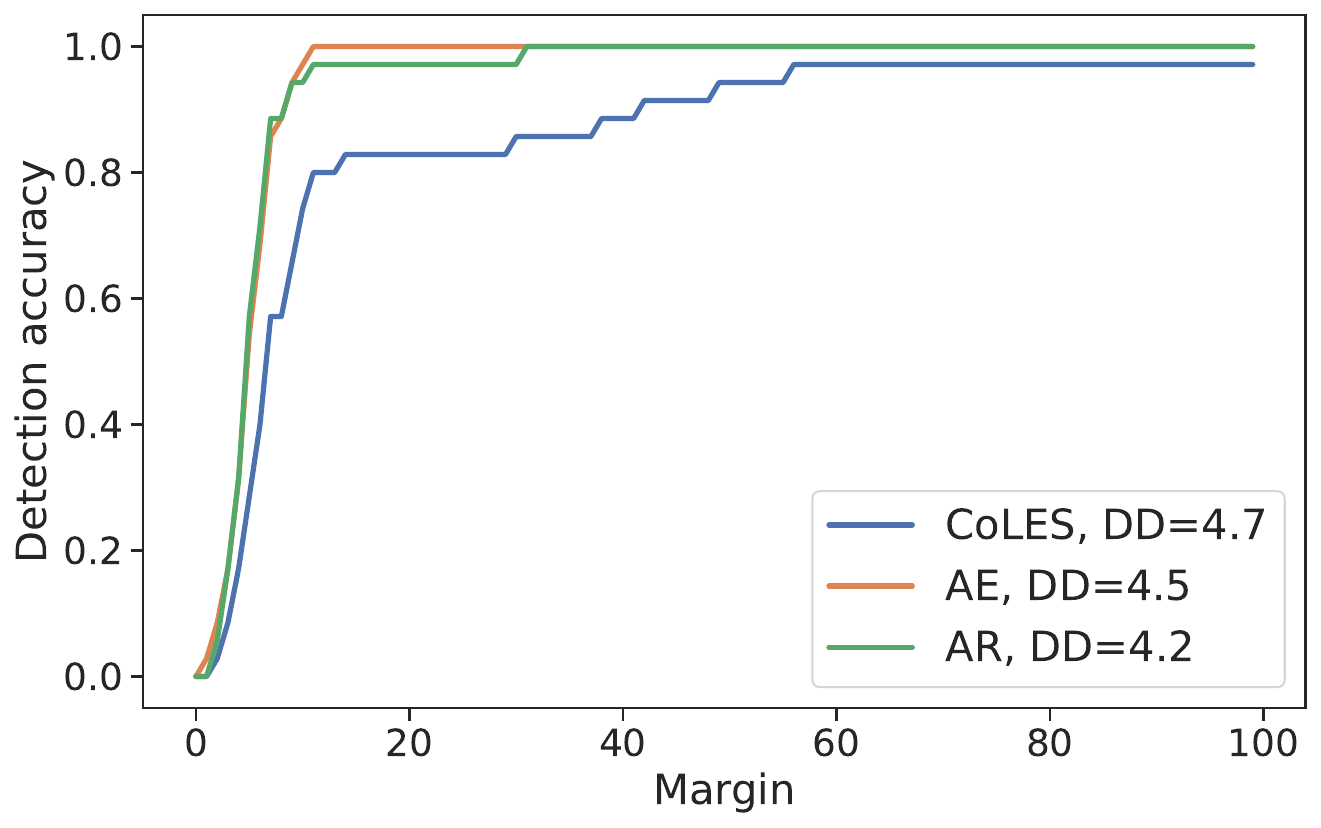}
\captionsetup{justification=centering, width=0.9\linewidth}
\captionof{figure}{Dependency of CPD accuracy on the margin $m$. Mean detection delays for each model are provided in the legend. We want to maximize detection accuracy and minimize detection delay.}
\label{fig:currency_change_point}
\end{figure}

We build local representations of the above data via pre-trained models and run a Change Point Detection (CPD) procedure on top of them. The detection accuracy and detection delay (two important metrics in the CPD domain) are shown in Figure~\ref{fig:currency_change_point}. 

AE and AR provide higher detection accuracy than CoLES for small margins. Moreover, the use of AR embeddings results in the lowest detection delays. It can be concluded that generative models are more sensitive to local pattern changes than contrastive ones.

\subsection{Result on improving embeddings via external information}
\label{sec:external_quality}

In this section, we describe the results of experiments aimed at obtaining an external context representation and using it to improve existing models.

The context vector is built on embeddings from a pre-trained CoLES encoder. To this end, we investigate various types of representation aggregations: averaging (\textit{Mean}), maximization (\textit{Max}), and an attention mechanism, with or without a trainable matrix (\textit{Learnable attention} and \textit{Attention}). The results are compared with a conventional CoLES encoder without adding external information (\textit{No pooling}). 

\begin{figure}[!ht]
     \centering
     \includegraphics[width=\textwidth]{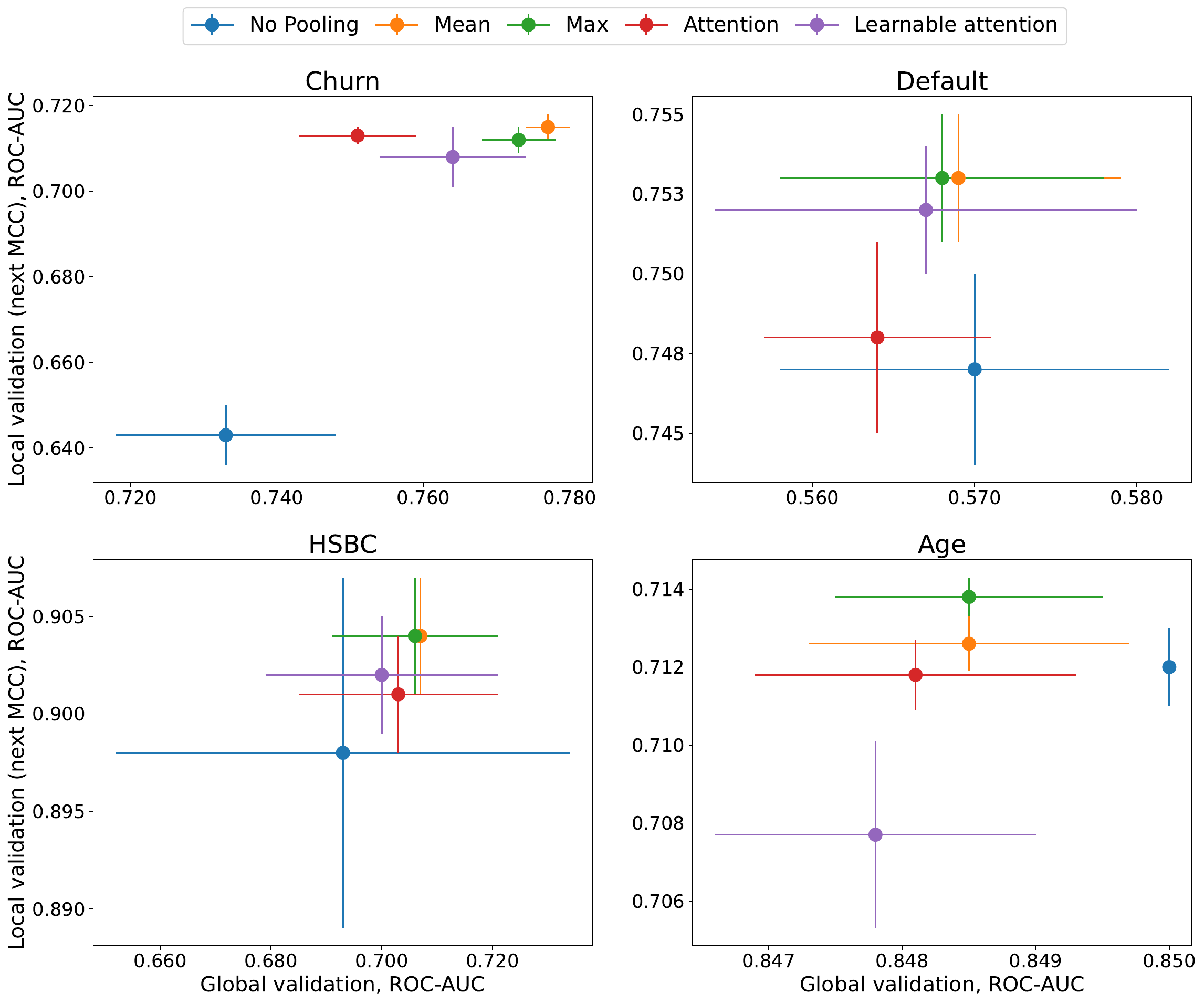}
     \caption{Quality of the various types of aggregation regarding their global and local properties. The x-axis at each graph corresponds to global validation ROC-AUC, while the y-axis shows the next MCC prediction ROC-AUC. Thus, the upper and righter the dot is, the better model it represents.}
     \label{fig:local_vs_global_pooling_quality}
\end{figure}

As in previous experiments, we demonstrate our results in two ways. Quantitative metrics are presented in Table~\ref{tab:external_table}, while qualitative trade-offs between global and local tasks are shown in Figure~\ref{fig:local_vs_global_pooling_quality}. All results are averaged across five different pre-trained encoder models.

\begin{table}[!ht]
\begin{adjustbox}{width=\textwidth}
\begin{tabular}{l|ccc|ccc|ccc}
\hline
\multirow{2}{*}{Method} & 
\multicolumn{3}{c|}{Global validation} & 
\multicolumn{3}{c|}{Local validation: next MCC} & 
\multicolumn{3}{c}{Local validation: binary target} \\ \cline{2-10}
& Accuracy & ROC-AUC & PR-AUC & Accuracy & ROC-AUC & PR-AUC & Accuracy & ROC-AUC & PR-AUC \\ \hline
\multicolumn{10}{c}{Churn} \\ \hline
No pooling & 0.67\scriptsize{$\pm$0.02} & 0.73\scriptsize{$\pm$0.02} & 0.77\scriptsize{$\pm$0.03} & 0.23\scriptsize{$\pm$0.00} & 0.64\scriptsize{$\pm$0.01} & 0.16\scriptsize{$\pm$0.01}	& \textbf{0.73}\scriptsize{$\pm$0.00} & 0.56\scriptsize{$\pm$0.01} & 0.32\scriptsize{$\pm$0.01} \\ \hdashline
Mean & \textbf{0.71}\scriptsize{$\pm$0.01} & \textbf{0.78}\scriptsize{$\pm$0.00} & \textbf{0.83}\scriptsize{$\pm$0.00}	& 0.26\scriptsize{$\pm$0.01} & \textbf{0.72}\scriptsize{$\pm$0.00}	& \textbf{0.21}\scriptsize{$\pm$0.00}	& \textbf{0.73}\scriptsize{$\pm$0.00} & 0.59\scriptsize{$\pm$0.01} & 0.35\scriptsize{$\pm$0.01} \\
Max & \textbf{0.71}\scriptsize{$\pm$0.01} & 0.77\scriptsize{$\pm$0.01} & 0.82\scriptsize{$\pm$0.01}	& 0.26\scriptsize{$\pm$0.01} & 0.71\scriptsize{$\pm$0.00}	& \textbf{0.21}\scriptsize{$\pm$0.00}	& \textbf{0.73}\scriptsize{$\pm$0.00}	& 0.65\scriptsize{$\pm$0.03} & 0.40\scriptsize{$\pm$0.02} \\
Attention & 0.69\scriptsize{$\pm$0.01} & 0.75\scriptsize{$\pm$0.01} & 0.80\scriptsize{$\pm$0.01} & \textbf{0.27}\scriptsize{$\pm$0.01} & 0.71\scriptsize{$\pm$0.00} & \textbf{0.21}\scriptsize{$\pm$0.00} & \textbf{0.73}\scriptsize{$\pm$0.00}	& 0.60\scriptsize{$\pm$0.00} & 0.34\scriptsize{$\pm$0.01} \\
Learn. attention & 0.70\scriptsize{$\pm$0.01} & 0.76\scriptsize{$\pm$0.01} & 0.82\scriptsize{$\pm$0.00}	& 0.26\scriptsize{$\pm$0.01} & 0.71\scriptsize{$\pm$0.01}	& \textbf{0.21}\scriptsize{$\pm$0.00}	& 0.71\scriptsize{$\pm$0.00} & \textbf{0.67}\scriptsize{$\pm$0.00}	& \textbf{0.43}\scriptsize{$\pm$0.01} \\ \hline
\multicolumn{10}{c}{Default} \\ \hline
No pooling & 0.96\scriptsize{$\pm$0.00}	& \textbf{0.57}\scriptsize{$\pm$0.01} & 0.06\scriptsize{$\pm$0.01} & 0.34\scriptsize{$\pm$0.01} & 0.75\scriptsize{$\pm$0.00} & 0.26\scriptsize{$\pm$ 0.00}	& 0.99\scriptsize{$\pm$0.00} & 0.54\scriptsize{$\pm$0.04} & 0.01\scriptsize{$\pm$0.00} \\ \hdashline
Mean & 0.96\scriptsize{$\pm$0.00} & \textbf{0.57}\scriptsize{$\pm$0.01}	& 0.06\scriptsize{$\pm$0.01} & 0.34\scriptsize{$\pm$0.01} & 0.75\scriptsize{$\pm$0.00} & \textbf{0.27}\scriptsize{$\pm$0.00} & 0.99\scriptsize{$\pm$0.00} & \textbf{0.65}\scriptsize{$\pm$0.06} & 0.01\scriptsize{$\pm$0.00} \\
Max	& 0.96\scriptsize{$\pm$0.00} & \textbf{0.57}\scriptsize{$\pm$0.01}	& 0.06\scriptsize{$\pm$0.01} & 0.34\scriptsize{$\pm$0.01} & 0.75\scriptsize{$\pm$0.00} & \textbf{0.27}\scriptsize{$\pm$0.00} & 0.99\scriptsize{$\pm$0.00} & 0.44\scriptsize{$\pm$0.11} & 0.01\scriptsize{$\pm$0.01} \\
Attention & 0.96\scriptsize{$\pm$0.00} & 0.56\scriptsize{$\pm$0.01}	& 0.06\scriptsize{$\pm$0.01} & 0.34\scriptsize{$\pm$0.01} & 0.75\scriptsize{$\pm$0.00} & 0.26\scriptsize{$\pm$0.00} & 0.99\scriptsize{$\pm$0.00} & 0.61\scriptsize{$\pm$0.06} & 0.01\scriptsize{$\pm$0.00} \\
Learn. attention & 0.96\scriptsize{$\pm$0.00} & \textbf{0.57}\scriptsize{$\pm$0.01} & 0.06\scriptsize{$\pm$0.01} & 0.34\scriptsize{$\pm$0.01} & 0.75\scriptsize{$\pm$0.00} & \textbf{0.27}\scriptsize{$\pm$0.00} & 0.99\scriptsize{$\pm$0.00} & 0.64\scriptsize{$\pm$0.09} & 0.01\scriptsize{$\pm$0.00} \\ \hline
\multicolumn{10}{c}{HSBC} \\ \hline
No pooling & 0.92\scriptsize{$\pm$0.00} & 0.69\scriptsize{$\pm$0.04} & 0.15\scriptsize{$\pm$0.02} & 0.69\scriptsize{$\pm$0.02} & 0.90\scriptsize{$\pm$0.01} & 0.82\scriptsize{$\pm$0.01} & 1.00\scriptsize{$\pm$0.00} & 0.38\scriptsize{$\pm$0.09} & 0.01\scriptsize{$\pm$0.00} \\ \hdashline
Mean & 0.92\scriptsize{$\pm$0.00} & \textbf{0.71}\scriptsize{$\pm$0.01} & \textbf{0.25}\scriptsize{$\pm$0.01}	& 0.69\scriptsize{$\pm$0.02} & 0.90\scriptsize{$\pm$0.00} & \textbf{0.83}\scriptsize{$\pm$0.01} & 1.00\scriptsize{$\pm$0.00} & 0.39\scriptsize{$\pm$0.21} & 0.01\scriptsize{$\pm$0.01} \\
Max	& 0.92\scriptsize{$\pm$0.00} & \textbf{0.71}\scriptsize{$\pm$0.02}	& 0.21\scriptsize{$\pm$0.02} & 0.70\scriptsize{$\pm$0.02} & 0.90\scriptsize{$\pm$0.00} & \textbf{0.83}\scriptsize{$\pm$0.01} & 1.00\scriptsize{$\pm$0.00} & 0.41\scriptsize{$\pm$0.19} & 0.01\scriptsize{$\pm$0.00} \\
Attention & 0.92\scriptsize{$\pm$0.00} & 0.70\scriptsize{$\pm$0.02} & 0.18\scriptsize{$\pm$0.01} & \textbf{0.74}\scriptsize{$\pm$0.02} & 0.90\scriptsize{$\pm$0.00} & \textbf{0.83}\scriptsize{$\pm$0.01} & 1.00\scriptsize{$\pm$0.00} & \textbf{0.47}\scriptsize{$\pm$0.20} & 0.01\scriptsize{$\pm$0.00} \\
Learn. attention & 0.92\scriptsize{$\pm$0.00} & 0.70\scriptsize{$\pm$0.02} & 0.22\scriptsize{$\pm$0.02}	& 0.70\scriptsize{$\pm$0.02} & 0.90\scriptsize{$\pm$0.00} & \textbf{0.83}\scriptsize{$\pm$0.01} & 1.00\scriptsize{$\pm$0.00} & 0.38\scriptsize{$\pm$0.14} & 0.01\scriptsize{$\pm$0.00} \\ \hline
\multicolumn{10}{c}{Age} \\ \hline
No pooling & 0.61\scriptsize{$\pm$0.00} & 0.85\scriptsize{$\pm$0.00} & \textbf{0.66}\scriptsize{$\pm$0.00} & 0.32\scriptsize{$\pm$0.00} & 0.71\scriptsize{$\pm$0.00} & 0.24\scriptsize{$\pm$0.00} & ~ & ~ & ~ \\ \cdashline{1-7}

Mean & \textbf{0.64}\scriptsize{$\pm$0.03} & 0.85\scriptsize{$\pm$0.00} & 0.63\scriptsize{$\pm$0.02} & \textbf{0.33}\scriptsize{$\pm$0.00} & 0.71\scriptsize{$\pm$0.00} & 0.24\scriptsize{$\pm$0.00} & ~ & ~ & ~ \\

Max & \textbf{0.64}\scriptsize{$\pm$0.03} & 0.85\scriptsize{$\pm$0.00} & 0.63\scriptsize{$\pm$0.02} & \textbf{0.33}\scriptsize{$\pm$0.00} & 0.71\scriptsize{$\pm$0.00} & 0.24\scriptsize{$\pm$0.00} & ~ & NA & ~ \\

Attention & \textbf{0.64}\scriptsize{$\pm$0.03} & 0.85\scriptsize{$\pm$0.00} & 0.63\scriptsize{$\pm$0.02} & \textbf{0.33}\scriptsize{$\pm$0.00} & 0.71\scriptsize{$\pm$0.00} & 0.24\scriptsize{$\pm$0.00} &  ~ & ~ & ~ \\

Learn. attention & 0.61\scriptsize{$\pm$0.01} & 0.85\scriptsize{$\pm$0.00} & 0.65\scriptsize{$\pm$0.00} & \textbf{0.33}\scriptsize{$\pm$0.00} & 0.71\scriptsize{$\pm$0.00} & 0.24\scriptsize{$\pm$0.00} & ~ & ~ & ~ \\ \hline
\end{tabular}
\end{adjustbox}
\caption{
Quality metrics for global and local embedding validation results for external information addition. 
All metrics in the Table should be maximized. 
The results are averaged by five runs and are given in the format $mean \pm std$. 
The best values are \textbf{highlighted}.
}
\label{tab:external_table}
\end{table}

The experimental results indicate that using external context generally leads to improved performance. This is especially evident in the case of the balanced Churn dataset, where improvements can be noticed in all testing scenarios. 

The Attention and Learnable attention approaches are often among the most effective, or at least comparable to the best approaches, especially in local validation tasks. It meets our expectations as the attention mechanism is known to assist in identifying local patterns within sequences (see~\ref{sec:global_context_methods}).

The Mean and Max methods are often the top-1 and top-2 performers in global validation tasks, respectively. This behavior is also explainable, as both approaches smooth local patterns while aggregating representations. While this may be beneficial for global properties, it negatively affects the quality of local ones.

\subsection{Ranking and Discussion}\label{chap:discussion}
In this research, we compare several classes of transactional data models in terms of their embeddings' global and local properties. 
This section attempts to summarize these findings by in-depth analysis and supplementary ranking with respect to ROC-AUC values for each task.

The rankings compare all the models from~\autoref{tab:main_results}, as well as "CoLES ext." -- the CoLES model, modified by the mean external information inclusion procedure, which is a reasonable compromise judging by~\autoref{tab:external_table}.
This is more illustrative than the specific metric values from~\autoref{tab:main_results} since here we equate the statistically indistinguishable results.
We increase the rank if the p-value is less than $0.1$ for a one-sided T-test, i.e. only if the increment is large enough compared to the corresponding variances.

\subsubsection{Event type.}
The resulting per-dataset ranks for the local event type task are presented in \autoref{tab:event_type_ranks}.
As stated above, the autoregressive model is a clear winner here: this indicates a robustly superior ability to capture local patterns and use them for forecasting.

The MLM approach makes a close second, dropping to third place only on the Age dataset, where it also gives in to the AE model.
This can be explained by a greater dependence on the order in which transactions are made: the Transformer model used in MLM is known for ignoring positional information (see e.g.~\cite{haviv2022transformer}), while the RNN used in AE takes it into account.

Next, interestingly enough, CoLES with external information slightly outperforms the AE model on average.
It seems that the external information gives the edge necessary for the advanced CoLES method to sometimes beat a simple but specialized autoencoder on local tasks, providing statistics on how other clients differ from each other at a specific moment in time.
Overall, as we stated previously, generative methods outperform vanilla contrastive methods, with the latter performing on-par with supervised baselines.

The ranking concludes with the temporal point process baselines.
This agrees with prior research: authors of~\cite{takimoto2024meta} state that TPP methods require long sequences to operate successfully, which is not the case for the local task.

\newcommand{\ctrm}[1]{{\color{OliveGreen} #1}\textsuperscript{\dag}}
\newcommand{\tppm}[1]{{\color{Fuchsia} #1}\textsuperscript{\ddag}}
\newcommand{\genm}[1]{{\color{RoyalBlue} #1}\textsuperscript{\S}}

\begin{table}[!ht]
    \centering
    \caption{Model ranking for the local event type prediction task. Equal ranks correspond to indistinguishable performance. Models are colour-coded with superscripts for monochrome versions: \genm{blue} for generative, \ctrm{green} for contrastive and \tppm{fuchsia} for TPP.}
    \begin{tabular}{lccccc}
    \hline
        ~ & Age & Churn & Default & HSBC & Mean \\ \hline
        \genm{AR} & 1 & 1 & 1 & 1 & 1.00 \\ 
        \genm{MLM} & 3 & 1 & 2 & 1 & 1.75 \\ 
        \ctrm{CoLES ext.} & 3 & 2 & 2 & 3 & 2.50 \\ 
        \genm{AE} & 2 & 3 & 4 & 2 & 2.75 \\ 
        \ctrm{CoLES} & 3 & 4 & 3 & 3 & 3.25 \\ 
        Best baseline & 4 & 4 & 5 & 3 & 4.00 \\ 
        \ctrm{TS2Vec} & 6 & 4 & 5 & 4 & 4.75 \\ 
        \tppm{A-NHP} & 5 & 5 & 6 & 4 & 5.00 \\ 
        \tppm{NHP} & 5 & 5 & 6 & 4 & 5.00 \\ 
        \tppm{COTIC} & 6 & 6 & 7 & 5 & 6.00 \\ \hline
    \end{tabular}
    \label{tab:event_type_ranks}
\end{table}

\subsubsection{Local binary.}
The ranks for the local binary task are presented in~\autoref{tab:local_binary_ranks}.
Overall, contrary to the previously considered event-type prediction task, the generative methods now lag behind the modified CoLES method regarding the mean model rank.
Specifically, the generative methods show poor performance on the Default dataset.
This can be explained by the label's more global nature: after all, it was deduced from a sequence-wise, global label via cropping, and the probability of loan repayment likely changes slowly with time.
Moreover, the highly-local AR model comes in fourth here, which further supports this claim.
Another very important factor here is class imbalance.
Remember, the Default dataset is highly imbalanced by itself; our local label extraction procedure only makes matters worse.
It can be concluded, that contrastive and TPP models handle such extreme class asymmetry better, able to extract rare patterns, while generative models focus mostly on statistically likely ones.

The Churn dataset does not distinguish much between generative methods and "CoLES ext.": the label here seems to require both local and global properties.
Lastly, the HSBC dataset, which initially had a local label (indicating specifically which transactions are fraudulent), favours generative models, which is in line with our reasoning.

Finally, it can be observed that COTIC and TS2Vec perform surprisingly well on HSBC.
Both of these models use convolutions.
Good performance may be due to the fact that they are able to meticulously analyse short, fixed-length sequences, paying attention (via max pooling) to the most suspicious transaction.

\begin{table}[!ht]
    \centering
    \caption{Model ranking for the local binary label prediction task. Equal ranks correspond to indistinguishable performance. Models are colour-coded with superscripts for monochrome versions: \genm{blue} for generative, \ctrm{green} for contrastive and \tppm{fuchsia} for TPP.}
    \begin{tabular}{lcccc}
    \hline
        ~ & Churn & Default & HSBC & Mean \\ \hline
        \ctrm{CoLES ext.} & 1 & 1 & 3 & 1.67 \\ 
        \genm{MLM} & 2 & 2 & 1 & 1.67 \\ 
        \genm{AR} & 1 & 4 & 1 & 2.00 \\ 
        \genm{AE} & 2 & 2 & 3 & 2.33 \\ 
        \ctrm{TS2Vec} & 4 & 1 & 2 & 2.33 \\ 
        \tppm{A-NHP} & 3 & 1 & 4 & 2.67 \\ 
        \tppm{NHP} & 3 & 2 & 3 & 2.67 \\ 
        \ctrm{CoLES} & 3 & 2 & 4 & 3.00 \\ 
        Best baseline & 3 & 3 & 4 & 3.33 \\ 
        \tppm{COTIC} & 4 & 5 & 2 & 3.67 \\ \hline
    \end{tabular}
    \label{tab:local_binary_ranks}
\end{table}

\subsubsection{Global.}
The ranks for the global task are presented in~\autoref{tab:global_ranks}.
It comes without surprise that the CoLES variations take the lead here, with the external information model showing more stable performance than the vanilla option.
Specifically, the plain CoLES lags behind on the Churn task.
The final dynamics right before a client leaves the bank are presumably very important here, making Churn somewhat local, and, as mentioned previously, adding external information improves CoLES's performance on local tasks.
This point is supported by the fact that the contrastive TS2Vec method also performs poorly on the Churn task, while the local generative methods show good results here.
A more interesting observation is that TPP methods rise up to a respectable second place on Churn: a client's transactions grow less frequent before he leaves the bank, which makes this task very well suited for these time-sensitive models.

Next come the generative models, which mostly show good results here.
A notable exception is provided by our best event type model, AR: it loses to MLM, AE, and even the supervised baseline for the HSBC task.
It seems that its autoregressive nature renders it too local. It has a very narrow view of only the few last transactions, making it forget fraudulent ones that may have appeared early on in the sequence.

The ranking concludes with TPP methods, showing highly unstable performance between datasets.
All these models take into the account the temporal structure of the sequence, which appears to be a dead weight for the global task, only hindering their learning.

This table presents several anomalies.
For one, the supervised baseline is ranked first for the Age task.
As mentioned above, the best way to handle Age is to look for certain locally ordered patterns within the sequence, which is exactly what a simple RNN does.
Since the sequences are long, their lengths vary little (in a narrow range from 700 to a little over 1k elements) and there is plenty of labelled data, the baseline is able to accumulate enough information to tune a good, specialized classifier.

Next, the NHP model achieves high performance on the Churn task.
As we have concluded from the local binary label, churning may be predictable based on the time intervals closer to the end of the sequence.
NHP appears to capture these dynamics well, without introducing restrictive complex modifications.

Finally, COTIC performs on par with the external information CoLES on Default.
This supports our hypothesis that the continuous-time convolutions work well on highly regular sequence sizes (all sequences from default are of length 300), extracting local and global patterns from fixed positions in the sequence and combining them into highly informative representations.

\begin{table}[!ht]
    \centering
    \caption{Model ranking for the global label prediction task. Equal ranks correspond to indistinguishable performance. Models are colour-coded with superscripts for monochrome versions: \genm{blue} for generative, \ctrm{green} for contrastive and \tppm{fuchsia} for TPP.}
    \begin{tabular}{lccccc}
    \hline
        ~ & Age & Churn & Default & HSBC & Mean \\ \hline
        \ctrm{CoLES ext.} & 1 & 1 & 1 & 1 & 1.00 \\ 
        \ctrm{CoLES} & 1 & 3 & 1 & 1 & 1.50 \\ 
        \genm{MLM} & 2 & 2 & 2 & 2 & 2.00 \\ 
        Best baseline & 1 & 4 & 2 & 2 & 2.25 \\ 
        \genm{AR} & 3 & 2 & 1 & 3 & 2.25 \\ 
        \genm{AE} & 4 & 2 & 2 & 2 & 2.50 \\ 
        \tppm{NHP} & 5 & 2 & 2 & 3 & 3.00 \\ 
        \tppm{COTIC} & 6 & 3 & 1 & 4 & 3.50 \\ 
        \ctrm{TS2Vec} & 2 & 5 & 2 & 5 & 3.50 \\ 
        \tppm{A-NHP} & 5 & 3 & 3 & 4 & 3.75 \\ \hline
    \end{tabular}
    \label{tab:global_ranks}
\end{table}

\subsubsection{Executive Summary}
Now, we will briefly summarize the above discussion in a few points.
Overall, the generative methods clearly lead at the event-type task (since they were effectively trained for it).
The local binary and global tasks offer fine-grained insights, based on the label's nature.
\begin{itemize}
    \item The Age dataset has regular, simple, ordered patterns, appearing all throughout the sequence, indicative of the final label. 
    The event-type task is thus best solved by ordered generative models (using RNN as the backbone), and the global one, having enough transactions and data -- by the supervised baseline.
    \item The Churn label is largely defined by the time intervals closer to the end of the sequence.
    Consequently, this task requires a compromise between local and global properties, best achieved by CoLES with the inclusion of extrnal information.
    The local binary task, although devised from a global label, also relies greatly on fine-grained dynamics.
    The AR performs well here, on par with "CoLES ext.".
    \item The HSBC dataset has a truly local binary label, which makes it well-suited for generative models.
    However, the continuous-convolution approaches are also a great fit for this task: the cropped sequence length is fixed during testing, so they are able to pay attention to the most suspicious transactions via max-pooling.
    On the other hand, the global task requires broad attention to whole sequences of different lengths, favouring contrastive approaches over generative and TPP methods.
    \item The Default dataset has proven to be the hardest for all the considered models, without any clear winners on downstream tasks.
    The local binary label seems to favour contrastive and TPP models: we argue this is due to the label leaning more on global properties, as well as extreme class imbalance.
    Unsurprisingly, both CoLES methods take first place in the global ranking.
    However, the victory is shared with continuous convolution methods: the sequences here are also of fixed size, both during testing and training, which appears to be the speciality of these approaches.
\end{itemize}

%% file: 7-conclusions.tex
This paper examines various methods for extracting meaningful representations from transaction data, evaluating their effectiveness in capturing both global and local patterns of transactional sequences. 
We adapted generative self-supervised learning methodologies (specifically AE, MLM, and AR models) to the domain of transaction sequences, resulting in embeddings that effectively capture local and dynamic features, while reasonably addressing global patterns inherent in transactional data. 
These models were rigorously compared against state-of-the-art domain-specific deep learning techniques, including contrastive self-supervised learning (such as CoLES and TS2Vec), temporal point process models (HNP, A-NHP, and COTIC), and conventional supervised learning approaches.

Our comprehensive benchmark, i.e. one that simultaneously addresses all desired representation properties, reveals that no single method is universally superior for all applied problems.
Contrastive approaches generally excel in global classification tasks, making them effective for situations requiring a broader view of transactional data. 
In contrast, generative models are better suited for capturing local patterns, essential for scenarios involving immediate customer interactions and behavior prediction. 
Temporal point process models, which are typically designed for process intensity restoration, tend to underperform in representation learning contexts.
However, their ability to account for time intervals makes them a great choice for certain scenarios.
Finally, the convolution models (TS2Vec, COTIC) excel at uniform-length sequences.

Among the classic methods evaluated, the proposed AR model stands out for its robust performance across various tasks. 
It slightly gives way to CoLES, yet it remains comparably effective in global classification while surpassing all other models in next MCC code prediction tasks, indicating its ability to effectively balance the recognition of both global and local patterns within transactional data.

Additionally, we introduced an innovative method for incorporating external contextual information into transactional representations. 
By leveraging the activity of other clients, this method constructs an external context vector, significantly improving model performance, with an increase in ROC-AUC of up to 20\% for local tasks, making it the clear winner in our downstream benchmarks. 
The best results were achieved using a trainable attention mechanism that accounts for embeddings of similar clients, however the simpler mean aggregation also gives a comparable boost in metrics, making it a reasonable compromise.

Overall, our research provides a benchmark for evaluating diverse properties of various representation learning methods applied to financial transaction data. 
These techniques offer scalable and adaptable solutions that can be applied across a wide range of banking applications, enhancing decision-making and customer experience. 
The contributions made to representation learning in this paper have significant potential for advancing the banking industry. 
By effectively utilizing AI, banks can optimize operations, reduce risks, and offer more personalized services to their clients. 
As AI continues to develop, these findings and methods will drive further innovation in financial services.

%% file: 8-appendix.tex
\section{Implementation details}
\label{sec:implementation_details}

\paragraph{\textbf{Main models}}
In this research, we generally follow the pipeline of the \href{https://github.com/dllllb/pytorch-lifestream/tree/main}{pytorch-lifestream} Python package, which contains the implementation of the CoLES model.
Additionally, we use NHP and A-NHP models from the EasyTPP~\cite{xue2023easytpp} library available \href{https://github.com/ant-research/EasyTemporalPointProcess/tree/main}{online}.
Finally, the COTIC model implementation was taken from the original GitHub \href{https://github.com/VladislavZh/COTIC/tree/main}{repository} and adopted into our pipeline.

Information about the architectures and hyperparameters of the models used in our study is given below.

\begin{itemize}
    \item CoLES is a one-layer recurrent neural network. 
    We use LSTM with a hidden layer dimension of $1024$ for the Churn, Age, and HSBC datasets and $512$ for industrial data. 
    For Default, we select GRU with a hidden size of $800$.
    The models were trained for $60$ epochs with a batch size of $128$.

    \item AE's encoder and AR have identical architectures for all the public datasets ---  LSTM with a hidden layer dimension of $1024$.
    The hidden size of $512$ was used for industrial data.
    AR was trained for $60$ epochs with a batch size of $128$.

    \item The AE decoder is an LSTM with a hidden layer dimension twice bigger than in the encoder, 1024 for industrial data and 2048 for other datasets. The model was trained for $2000$ steps with a batch size of $1024$.
    
    \item The MLM is a Transformer with six hidden layers with eight heads; the embedding dimension and the perceptron hidden layer dimension equals $512$ for industrial data and $1024$ for other datasets. 
    The training regimen is similar to that of the AE's decoder: $2000$ steps on batches of $1024$.

    \item For AR, MLM, and AE models, we use a combination of two losses described in~\ref{sec:ae}. The weights for these losses are set to five and one, respectively.
    
    \item TS2Vec has ten one-dimensional convolutional blocks with an expanded kernel size of three, an expansion factor of $2^l$ (where $l$ is the block number), and the number of output channels equals $512$ for private data and $1024$ for other datasets.
    It was trained for $100$ epochs with batches of $1024$.

    \item The COTIC model has five continuous one-dimensional convolutional layers with a hidden dimension of $32$ and a kernel size of five. 
    Its kernel is a multi-layer perceptron with hidden layer dimensions eight, four, and eight, and the prediction head is a two-layer perceptron with a hidden dimension of $100$.
    The model was trained for $100$ epochs with a batch size of $20$. 
    We had to significantly limit the number of parameters to reduce the memory consumption and the training time since COTIC is inherently less efficient than other models.
    
    \item NHP has a custom continuous-time LSTM architecture with exponential decay of hidden states as in original work ~\cite{mei2017neural}.
    We mainly adopt the parameters from the EasyTPP experiment configuration files while increasing the hidden dimension of the model up to $64$ to be comparable with the other TPP models.
    It was trained for $50$ epochs with a batch size of $128$.
    
    \item The encoder of the A-NHP model consists of two multi-head attention blocks with a final embedding dimension of $64$.
    It also uses a temporal encoding with a time embedding size of four.
    The model was trained for $50$ epochs with batches of $32$.
\end{itemize}

\paragraph{\textbf{Input features}}

Note that all self-supervised approaches (both contrastive and generative) take two features as input: MCCs and transaction amounts. 
As MCC is a categorical variable, we use its embedding of size $24$ (for Churn and HSBC) or $16$ (for Age and Default).

In contrast, the considered TPP models (NHP, A-NHP, COTIC) do not process amounts but take times between events as a part of the input. 
Here, we cannot use raw UNIX timestamps due to numeric overflows. Thus, all the times are normalized to represent the number of days (float) since the first transaction in the dataset.
Furthermore, to foster the quality of next event type prediction, we clip the rare MCC codes as it is done for the autoencoder models~\ref{sec:ae}.

\paragraph{\textbf{Models with external context}}
A model using external information requires quite a lot of computational resources since it needs to store local representations of all users from the training set for all available time points (Subsection~\ref{sec:global_context_methods}).
Due to memory constraints, we store only a random subset of local representations: for the Churn dataset, the number of clients to train in all experiments was $500$; for the Default dataset --- $150$; for HSBC --- $300$; and for Age --- $2000$.
Additionally, the batch is reduced to eight. 
Other hyperparameters for the encoder remain the same as in the experiments with the regular CoLES model.

\paragraph{\textbf{Validation model parameters}}
For the local validation procedure outlined in Section~\ref{chap:method}, we used the following hyperparameter values: window size is $32$, shift step is $16$, batch size is $512$, and the maximum number of epochs is ten.